\documentclass[11pt]{article}
\usepackage{graphicx}
\usepackage{color}
\usepackage{amsmath}
\usepackage{amssymb}
\usepackage{amscd}
\usepackage{framed}
\usepackage{bbm}

\usepackage{fancyhdr,a4wide}
\usepackage{amsthm}

\newcommand{\R}{\mathbb{R}}
\newcommand{\inr}[1]{\left\langle #1 \right\rangle}

\newcommand{\E}{\mathbb{E}}
\newcommand{\M}{\mathbb{M}}
\newcommand{\Q}{\mathbb{Q}}
\newcommand{\B}{\mathbb{B}}
\newcommand{\HH}{\mathbb{H}}

\newcommand{\eps}{\varepsilon}

\newtheorem{Theorem}{Theorem}[section]
\newtheorem{Lemma}[Theorem]{Lemma}

\newtheorem{Definition}[Theorem]{Definition}

\newtheorem{Claim}[Theorem]{Claim}
\newtheorem{Corollary}[Theorem]{Corollary}
\newtheorem{Remark}[Theorem]{Remark}
\newtheorem{Example}[Theorem]{Example}
\newtheorem{Assumption}{Assumption}[section]

\numberwithin{equation}{section}

\def \proof {\noindent {\bf Proof.}\ \ }

\def \endproof
{{\mbox{}\nolinebreak\hfill\rule{2mm}{2mm}\par\medbreak}}
\def\IND{\mathbbm{1}}

\newcommand{\mP}{\mathcal{P}}

\def\IND{\mathbbm{1}}

\begin{document}
\title{An optimal unrestricted learning procedure}
\author{
Shahar Mendelson \thanks{Department of Mathematics, Technion, I.I.T, and Mathematical Sciences Institute, The Australian National University. Email: shahar@tx.technion.ac.il}}

\maketitle

\begin{abstract}
We study learning problems involving arbitrary classes of functions $F$, distributions $X$ and targets $Y$. Because \emph{proper} learning procedures, i.e., procedures that are only allowed to select functions in $F$, tend to perform poorly unless the problem satisfies some additional structural property (e.g., that $F$ is convex), we consider \emph{unrestricted learning procedures} that are free to choose functions outside the given class.

We present a new unrestricted procedure that is optimal in a very strong sense: the required sample complexity is essentially the best one can hope for, and the estimate holds for (almost) any problem, including heavy-tailed situations. Moreover, the sample complexity coincides with the what one would expect if $F$ were convex, even when $F$ is not. And if $F$ is convex, the procedure turns out to be proper. Thus, the unrestricted procedure is actually optimal in both realms, for convex classes as a proper procedure and for arbitrary classes as an unrestricted procedure.
\end{abstract}

\section{Learning procedures and unrestricted procedures}
In the standard setup in statistical learning theory, one is given a class of functions $F$ defined on a probability space $(\Omega,\mu)$. The goal is to identify, or at least mimic, a function in $F$ that is as close as possible to the unknown target random variable $Y$ in some appropriate sense. If $X$ is distributed according to $\mu$ then an obvious candidate for being considered ``as close as possible to $Y$ in $F$" is the function
$$
f^*={\rm argmin}_{f \in F} \E(f(X)-Y)^2;
$$
it minimizes the average cost (relative to the squared loss) one has to pay for predicting $f(x)$ instead of $y$. In a more geometric language, $f^*$ minimizes the $L_2$ distance between $Y$ and the class $F$, and in what follows we implicitly assume that such a minimizer exists.

What makes the learner's task a potentially difficult one is the limited information at his disposal: instead of knowing the distribution $X$ and the target random variable $Y$ (which would make identifying $f^*$ a problem in approximation theory), both $X$ and $Y$ are not known. Rather, the learner is given an independent sample $(X_i,Y_i)_{i=1}^N$, with each pair $(X_i,Y_i)$ distributed according to the joint distribution $(X,Y)$. Using the sample, the learner selects some function in $F$, hoping that it is almost as good a prediction of $Y$ as $f^*$ is. The selection is made via a \emph{learning procedure}, which is a mapping $\Phi:(\Omega \times \R)^N \to F$.

If $\tilde{f}=\Phi((X_i,Y_i)_{i=1}^N)$ is the selection made by the procedure given the data, its \emph{excess risk} is the conditional expectation
\begin{equation} \label{eq:prediction-error}
\E\left(\left(\tilde{f}(X)-Y\right)^2 | (X_i,Y_i)_{i=1}^N \right) - \E \left(f^*(X)-Y\right)^2 \equiv {\cal E}_p,
\end{equation}
and the procedure's success is measured through properties of  ${\cal E}_p$. Since the information the learner has is limited, it is unlikely that $\tilde{f}$ can always be a good guess, and therefore $\Phi$'s performance is measured using a probabilistic yardstick: the \emph{sample complexity}; that is, for a given accuracy $\eps$ and a confidence parameter $0<\delta<1$, the number of independent pairs $(X_i,Y_i)_{i=1}^N$ that are needed to ensure that
$$
{\cal E}_p \leq \eps \ \ {\rm with \ probability \ at \ least \ } 1-\delta.
$$

The key question in learning theory is to identify a procedure that performs with the optimal sample complexity (an elusive term that will be clarified in what follows) for each learning problem. It stands to reason that the optimal sample complexity should depend on the right notion of statistical complexity of the class $F$; on some (minimal) global information on the target $Y$ and underlying distribution $X$; and the required accuracy and confidence levels.

To put our results in context, let us begin by describing what we mean by optimal sample complexity and  optimal procedure. These are minor modifications of notions introduced in \cite{LugMen16}, in which an optimal proper learning procedure was identified for (almost) any problem involving a convex class $F$.

Before we dive into more technical details, let us fix some notation.

Throughout we denote absolute constants by $c$ or $C$. Their values may change from line to line. $C(\alpha)$ or $C_\alpha$ are constants that depend only on the parameter $\alpha$. $A \lesssim B$ means that there is an absolute constant $C$ such that $A \leq CB$ and $A \lesssim_\alpha B$ means that the constant depends only on the parameter $\alpha$. We write $A \sim B$ if $A \lesssim B$ and $B \lesssim A$, while $A \sim_\alpha B$ means that the equivalence constants depend only on $\alpha$.

All the functions we consider are square integrable on an appropriate probability space, though frequently we do not specify the space or the measure as those will be clear from the context. Thus, $\|f-h\|_{L_2}^2 = \E (f(X)-h(X))^2$ and $\|f-Y\|_{L_2}^2 = \|f(X)-Y\|_{L_2}^2=\E (f(X)-Y)^2$, and we adopt a similar notation for other $L_p$ spaces.

For the sake of simplicity, we denote each learning problem, consisting of the given class of functions $F$, an unknown underlying distribution $X$, and an unknown target $Y$, by the triplet $(F,X,Y)$. It should be stressed that when we write ``given the triplet $(F,X,Y)$", it does not mean that the learner has any additional information on $X$ or on $Y$. Still, this notation helps one to keep track of the fact that the sample complexity may change not only with $F$ but also with $X$ and $Y$.

We denote generic triplets by $(H,X,Y)$ and $(F,X,Y)$. For a triplet $(F,X,Y)$ we set  $f^*={\rm argmin}_{f \in F} \E (f(X)-Y)^2$ and $\sigma^2=\E(f^*(X)-Y)^2$. The class $F \pm H$ consists of all the functions $f \pm h$ for $f \in F$ and $h \in H$; also, for $\lambda \in [0,1]$, set $\lambda F =\{\lambda f : f \in F\}$.

\subsection{Notions of optimality}

The notion optimality we use is based on a list of `obstructions'. These obstructions, are, in some sense, trivial, and overcoming each one of them is something one would expect of any reasonable procedure---certainly from a procedure that deserves to be called optimal. On the other hand, overcoming each obstruction comes at a price: as we explain in what follows, a certain geometric obstruction forces one to consider procedures that need not be proper; and overcoming some trivial statistical obstructions requires a minimal number of sample points. 

\begin{framed}
Our main result shows that the sample size needed to overcome the trivial statistical obstructions suffices (up to some absolute multiplicative constant) for the solution of an (almost) arbitrary learning problem. And because of the geometric obstruction, the solution is carried out using an unrestricted procedure.
\end{framed}

Let us describe the `trivial' obstructions one may encounter and minimal price one has to pay to overcome each one.
\subsubsection*{A geometric obstruction}
In the standard ({\emph proper}) learning model the procedure is only allowed to take values in the given class $F$. At a first glance this restriction seems to be completely reasonable; after all, the learner's goal is to find a function that mimics the behaviour of the best function in $F$, and there is no apparent reason to look for such a function outside $F$. However, a more careful consideration shows that this restriction comes at a high cost:

\begin{Example} \label{ex:two-point-class}
Let $F=\{f_1,f_2\}$ and fix an integer $N$. Set $Y$ to be a `noisy' $\sim 1/\sqrt{N}$-perturbation of the midpoint $(f_1+f_2)/2$, that is slightly closer to $f_1$ than to $f_2$. Then, given samples $(X_i,Y_i)_{i=1}^N$, any proper procedure $\Phi$ will necessarily make the wrong choice with probability $1/10$; that is, with probability at least $1/10$,  $\Phi((X_i,Y_i)_{i=1}^N) = f_2$ and on that event, the excess risk is ${\cal E}_p \sim 1/\sqrt{N}$.
\end{Example}
In other words, by considering such targets, and given accuracy $\eps$, the sample complexity of any learning procedure taking values in $\{f_1,f_2\}$ cannot be better than $O(1/\eps^2)$ even if one is interested only in constant confidence.

A proof of this standard fact may be found, for example, in \cite{AnBa99,DGL96}.

\vskip0.3cm
Example \ref{ex:two-point-class} serves as a strong indication of a general phenomenon: there are seemingly simple problems, including ones involving classes with a finite number of functions (in this example, only two functions...), in which the sample complexity is significantly higher than what would be expected given the class' size. The reason for such \emph{slow rates} is that the `location' of the target relative to the class is not `favourable'\footnote{It is well understood that the target $Y$ is in a favourable location when the set of functions in $F$ that `almost minimize' the risk functional $f \to \E(f(X)-Y)^2$ consists only of perturbations of the unique minimizer $f^*$, see \cite{MR2426759} for more details.}. In contrast, if $F$ happens to be convex then any target is in a favourable location and there is no geometric obstruction that forces slow rates; the same also holds for a general class $F$ in the case of independent additive noise---when $Y=f_0(X)+W$ for some $f_0 \in F$ and $W$ that is mean-zero and independent of $X$.

Here we are interested in general triplets $(F,X,Y)$, making it is impossible to guarantee that the unknown target $Y$ is in a favourable location relative to $F$. Therefore, to have any hope of addressing this obstruction, one must remove the restriction that the procedure is proper; instead we consider \emph{unrestricted} procedures, that is, procedures that are allowed to take values outside $F$.

\subsubsection*{Statistical obstructions}
A natural way of finding generic statistical obstructions is identifying the reasons why a statistical procedure may make mistakes. Roughly put, there are two sources of error \cite{MenACM}:
\begin{description}
\item{$\bullet$} \emph{Intrinsic errors:} When $F$ is `rich' close to the true minimizer $f^*$, it is difficult to `separate' class members with the limited data the learner has. In noise-free (realizable) problems this corresponds to having a large \emph{version space}---the (random) subset of $F$, consisting of all the functions that agree with target on the given sample.
\item{$\bullet$} \emph{External errors:} When the `noise level' increases, that is, when $Y$ is relatively far form $F$, interactions between $Y$ and class members can cause distortions. These interactions make functions that are close to $f^*$ indistinguishable and causes the procedure to make mistakes in the choices it makes.
\end{description}
Obviously, describing the effect each one of these sources of error have on the sample complexity is of the utmost importance. The ``statistical obstructions" we refer to are defined for any class $F$ and underlying distribution $X$, and are the result of the intrinsic and external factors in two specific collections of learning problems involving $F$ and $X$ (keeping in mind that the learner does not know $X$). The targets $Y$ one considers are either:
\begin{description}
\item{$(1)$} \emph{Realizable targets}; that is, targets of the form $Y=f_0(X)$ where $f_0 \in F$; or
\item{$(2)$} \emph{Additive, independent gaussian noise}, that is, targets of the form $Y=f_0(X)+W$, where $f_0 \in F$ and $W$ is a centred gaussian random variable, independent of $X$ and with variance $\sigma^2$.
\end{description}

The idea is that an optimal statistical procedure must be able to address such simple problems, making them our choice of `trivial' statistical obstructions. And, the sample complexity needed to overcome the intrinsic and external errors for targets as in $(1)$ or $(2)$ is a rather minimal `price' one should be willing to pay when trying to address general prediction problems.

\vskip0.4cm

The first `trivial' statistical obstruction we consider has to do with realizable problems. Since the learner has no information on the underlying distribution $X$, there is no way of excluding the possibility that there are $f_1,f_2 \in F$ that are far from each other, and yet agree on a set of constant measure --- say $1/10$. Hence, given a sample of cardinality $N$, there is a probability of at least $\exp(-cN)$ that the two functions are indistinguishable on the sample. This trivial reason for having a version space with a large diameter sets the bar of the sample complexity at at-least $\sim \log(2/\delta)$.

\vskip0.4cm

The introduction of the other trivial obstructions requires additional notation. It is not surprising that the resulting sample complexity has to do with \emph{localized Rademacher averages}.

Let $D=\{f: \|f\|_{L_2}\le 1\}$ be the unit ball in $L_2(\mu)$, set $rD = \{f : \|f\|_{L_2} \leq r\}$ and put $S=\{f: \|f\|_{L_2}= 1\}$. The star-shaped hull of a class $F$ and a function $h$ is given by
$$
{\rm star}(F,h) = \{ \lambda f + (1-\lambda)h \ : \ 0 \leq \lambda
\leq 1, \ f \in F\};
$$
in other words, ${\rm star}(F,h)$ consists of the union of all the intervals whose end points are $h$ and $f \in F$. From here on we set
$$
F_{h,r} = {\rm star}(F-h,0) \cap rD = \left\{u=\lambda(f-h) : 0 \leq \lambda \leq 1, \ f \in F, \ \|u\|_{L_2} \leq r \right\}.
$$
Note that $F_{h,r}$ is the set one obtains by taking ${\rm star}(F,h)$, intersecting it with an $L_2(\mu)$ ball centred at $h$ and of radius $r$, and then shifting $h$ to $0$.


\begin{Definition} \label{def:trivial-complexity}
For a triplet $T=(F,X,Y)$ let
 \begin{equation} \label{eq:N-int}
N_{\rm int}(T,r,\kappa) =  \min \left\{ N : \E \sup_{u \in F_{f^*,r}} \left|\frac{1}{N}\sum_{i=1}^N \eps_i u(X_i) \right| \leq \kappa r \right\},
\end{equation}
and
\begin{equation} \label{eq:N-ext}
N_{\rm ext}(T,r,\kappa) = \min \left\{ N : \E \sup_{u \in F_{f^*,r}} \left|\frac{1}{N}\sum_{i=1}^N \eps_i (f^*(X_i)-Y_i)u(X_i) \right| \leq \kappa  r^2 \right\},
\end{equation}
where $(\eps_i)_{i=1}^N$ are independent, symmetric $\{-1,1\}$-valued random variables that are independent of $(X_i,Y_i)_{i=1}^N$ and the expectations are taken with respect to both $(\eps_i)_{i=1}^N$ and $(X_i,Y_i)_{i=1}^N$. 
\end{Definition}

Intuitively, $N_{\rm int}$ is the sample size needed to overcome `intrinsic' errors while $N_{\rm ext}$ is the sample size one must have to overcome the `external' ones. More accurately, one has the following:


\begin{Claim} \label{claim:lower} \cite{LecMen-subgauss16,MenLoc-v-Glob16}
There is an absolute constant $c_1$ for which the following holds. Under mild assumptions\footnote{The mild assumptions on $F$ and $X$ have to do with the continuity of the processes appearing in the definitions of $N_{\rm int}$ and $N_{\rm ext}$. We refer to \cite{LecMen-subgauss16,MenLoc-v-Glob16} for more information on these lower bounds.} on $F$ and $X$,
\begin{description}
\item{$\bullet$} If for every realizable target $Y=f_0(X)$, one has that with probability at least $3/4$ the diameter of the version space $\{f \in F : f(X_i)=Y_i \ \ {\rm for \ every \ }  1 \leq i \leq N\}$ is at most $\eps$, then the sample size is at least
    $$
    \sup N_{\rm int}(T,\sqrt{\eps},c_1),
    $$
where the supremum is taken with respect to all triplets $T=(F,X,Y)$ involving the fixed class $F$, the fixed (but unknown) distribution $X$, and targets of the form $Y=f_0(X)$, $f_0 \in F$.

\item{$\bullet$} If $\Phi$ is a learning procedure that performs with accuracy $\eps$ and confidence $3/4$ for any target of the form $Y=f_0(X)+W$ as in $(2)$, then it requires a sample size of cardinality at least
    $$
    \sup N_{\rm ext}(T,\sqrt{\eps},c_1),
    $$
    where the supremum is taken with respect to all triplets involving the fixed class $F$, the fixed (but unknown) distribution $X$, and targets of the form $Y=f_0(X)+W$, $f_0 \in F$.
\end{description}
\end{Claim}

Claim \ref{claim:lower} provides a lower bound on the sample complexity needed to overcome the trivial obstructions associated with $(1)$ and $(2)$ at a constant confidence level. When one is interested in a higher confidence level, one has the following:
\begin{Claim} \label{claim:lower-2} \cite{LecMen-subgauss16}
There is an absolute constant $c_2$ for which the following holds. Under mild assumptions on $F$ and $X$, any learning procedure $\Phi$ that performs with accuracy $\eps$ and confidence $1-\delta$ for any target of the form $Y=f_0(X)+W$ as in $(2)$, requires a sample size of cardinality at least
$$
c_2\frac{\sigma^2}{\eps} \cdot \log\left(\frac{2}{\delta}\right),
$$
where, as always, $\sigma=\|f^*(X)-Y\|_{L_2}$.
\end{Claim}

With the geometric obstruction and the trivial statistical obstructions in mind, a (seemingly wildly optimistic) notion of an optimal sample complexity and an optimal procedure is the following:
\begin{Definition} \label{def:optimal}
An unrestricted procedure is optimal if there are constants $c_1$ and $c_2$ such that for (almost) every triplet $T=(F,X,Y)$, the procedure performs with accuracy $\eps$ and confidence $1-\delta$ with sample complexity
\begin{equation} \label{eq:optimal-in-def}
N =  N_{\rm int}(T,\sqrt{\eps},c_1) + N_{\rm ext}(T,\sqrt{\eps},c_1) +c_2\left(\frac{\|f^*(X)-Y\|_{L_2}^2}{\eps}+1\right) \log\left(\frac{2}{\delta}\right).
\end{equation}
\end{Definition}

At a first glance, this benchmark seems to be too good to be true. One source of optimism is \cite{LugMen16} in which
a (proper) procedure that attains \eqref{eq:optimal-in-def} is established---the \emph{median-of-means tournament}. However, tournaments are shown to be optimal only for problems involving convex classes (or general classes but for targets that consist of independent additive noise). The success of the tournament procedure introduced in \cite{LugMen16} does not extend to more general learning problems; not only is it a proper procedure, its analysis uses the favourable location of the target in a strong way.

In what follows we build on ideas from \cite{LugMen16} and specifically on the notion of a median-of-means tournament and introduce a procedure that is essentially optimal in the sense of Definition \ref{def:optimal}.

\section{The main result in detail}
As a first step in an accurate formulation of our main result, let us specify what we mean by ``almost every triplet".

\begin{Assumption} \label{ass:main-on-F}
For a class $F$, let $U=(F+F)/2$ and assume that for every $0<\xi<1$ there exists $\kappa(\xi)$ such that for every $w \in U-U$,
\begin{equation} \label{eq:uniform-integrabilty-in-assumption}
\E w^2(X) \IND_{\{|w| \geq \kappa(\xi) \|w\|_{L_2}\}} \leq \xi \|w\|_{L_2}^2.
\end{equation}
\end{Assumption}

Equation \eqref{eq:uniform-integrabilty-in-assumption} is a uniform integrability condition for $U-U$, and as such it is only slightly stronger than a compactness assumption on $F$: \eqref{eq:uniform-integrabilty-in-assumption} holds for any $w \in L_2(\mu)$ individually, and the fact that $F$ is reasonably small allows the `cut-off' points $\kappa(\xi)$ to be chosen uniformly for any $w \in U-U$.



An indication that Assumption \ref{ass:main-on-F} is rather minimal is $L_q-L_2$ norm equivalence: that  there are constants $L$ and $q>2$ (which can be arbitrarily close to $2$) such that
$\|h\|_{L_q} \leq L \|h\|_{L_2}$ for every $h \in {\rm span}(F)$. An $L_q-L_2$ norm equivalence implies that Assumption \ref{ass:main-on-F} holds with $\kappa(\xi)$ depending only on $L$ and $q$, and the standard proof is based on tail integration and Chebychev's inequality.

Norm equivalence occurs frequently in statistical problems --- for example, in linear regression, where the class in question consists of linear functionals in $\R^d$.  It is standard to verify that $L_q-L_2$ norm equivalence is satisfied for random vectors $X$ that are subgaussian; log-concave; of the form $X=(x_i)_{i=1}^d$, where the $x_i$'s are independent copies of a symmetric, variance $1$ random variable that is bounded in $L_p$ for some $p \geq q$; and in many other situations (see, e.g. \cite{Men-General-loss17}).

\vskip0.3cm


\vskip0.3cm

While Assumption \ref{ass:main-on-F} is weaker than any $L_q-L_2$ norm equivalence, it is actually stronger than the small-ball condition which plays a central role in \cite{MenACM,Men-General-loss17,LugMen16}. Indeed, a small-ball condition means that there are $\gamma>0$ and $0<\delta<1$ such that
\begin{equation} \label{eq:sbc}
Pr(|w| \geq \gamma \|w\|_{L_2}) \geq \delta \ \ \ {\rm for \ any} \ w \in U-U.
\end{equation}
Invoking Assumption \ref{ass:main-on-F} for an arbitrary $0<\xi \leq 1/2$, it is evident that $w^\prime = w \IND_{\{|w| \leq \kappa(\xi) \|w\|_{L_2}\}}$ satisfies $\|w^\prime\|_{L_\infty}/\|w^\prime\|_{L_2} \leq \kappa(\xi)/(1-\xi)^{1/2}$ which, by the Paley-Zygmund Theorem, guarantees a small-ball condition for constants $\gamma$ and $\delta$ that depend only on $\xi$ and $\kappa(\xi)$.

The need for a slightly stronger assumption than \eqref{eq:sbc} arises because a small-ball condition leads only to an isomorphic lower bound on quadratic forms: it implies that with high probability,
\begin{equation} \label{eq:sbc-isomorphic}
\frac{1}{m}\sum_{i=1}^m w^2(X_i) \geq c \|w\|_{L_2}^2, \ \ \ {\rm for \ any} \ w \in U-U \ {\rm such \ that \ } \|w\|_{L_2} \geq r,
\end{equation}
but the constant $c$ cannot be made arbitrarily close to $1$. It turns out that proving that our procedure is optimal requires a version of \eqref{eq:sbc-isomorphic} for a constant that can be taken close to $1$.  We show in what follows that Assumption \ref{ass:main-on-F} suffices for that.

\vskip0.4cm

Next, we need an additional parameter that gives information on the way the target $Y$ interacts with the class $F$.
\begin{Definition} \label{def:main-on-Y}
For a triplet $T=(F,X,Y)$ set
$$
L_T = \sup_{f \in F} \left(\E \left(\frac{(f-f^*)(X)}{\|f-f^*\|_{L_2}}\right)^2 \cdot \left(\frac{f^*(X)-Y}{\sigma}\right)^2 \right)^{1/2}
$$
(recall that $\sigma=\|f^*(X)-Y\|_{L_2}$).

In particular, for any $f \in F$,
\begin{equation} \label{eq:L-Y-F}
\E (f-f^*)^2(X) \cdot (f^*(X)-Y)^2 \leq L_T^2 \sigma^2 \|f-f^*\|_{L_2}^2.
\end{equation}
\end{Definition}
Equation \eqref{eq:L-Y-F} plays a significant role in what follows, and the following examples may help in giving a better understanding of it:
\begin{description}
\item{$(1)$} If $Y=f_0(X)+W$ for some $f_0 \in F$ and $W$ is a mean-zero, square-integrable random variable that is independent of $X$ then $L_T=1$.
\item{$(2)$} Let $Y=f_0(X)+W$ for some $f_0 \in {\rm span}(F)$ and $W$ as in $(1)$. If for every $h \in {\rm span}(F)$, $\|h(X)\|_{L_4} \leq L \|h(X)\|_{L_2}$ then $L_T \leq L$. More generally, the same holds if the $L_4-L_2$ norm equivalence is true for $\E(Y|X)-f^*(X)$ and for every $(f-f^*)(X)$.
\item{$(3)$} If for every $f ,h\in F$, $\|f-h\|_{L_4} \leq L\|f-h\|_{L_2}$, then one may take $L_T=L(\|f^*(X)-Y\|_{L_4}/\sigma)$.
\end{description}
The proofs of all these observations is completely standard and we omit them.

\vskip0.4cm

Before we formulate the main result and for a reason that will become clear immediately, we need to outline a few preliminary details on the procedure we introduce. 

The procedure receives as input a class $H$ and a sample $(X_i,Y_i)_{i=1}^{2N}$, and returns a subset $H_1 \subset H$.
The two crucial features of $H_1$ are that
\begin{description}
\item{$\bullet$} It contains $h^*={\rm argmin}_{h \in H} \|h(X)-Y\|_{L_2}$; and
\item{$\bullet$} If $h \in H_1$ then either $h$ is `very close' to $h^*$ or alternatively, $(h+h^*)/2$ is much closer to $Y$ than $h^*$ is.
\end{description}

Now, let $T=(F,X,Y)$ be a triplet and fix an accuracy $\eps$ and a confidence level $\delta$. 
\begin{description}
\item{$\bullet$} Given an integer $N_1$, let $F_1$ is the set generated by procedure after being given the class $F$ and the sample $(X_i,Y_i)_{i=1}^{2N_1}$. 
\item{$\bullet$} Set $\bar{F}_1=(F_1+F_1)/2$ and let $\bar{T}_1$ be the triplet $(\bar{F}_1,X,Y)$. 
\item{$\bullet$} For an integer $N_2$ let $F_2$ to be the set generated by the procedure after being given $\bar{F}_1$ and an independent sample $(X_i,Y_i)_{i=1}^{2N_2}$. 
\item{$\bullet$} Let $\tilde{f}$ to be any function in $F_2$.
\end{description}


\begin{framed}
\begin{Theorem} \label{thm:main}
Let $(F,X,Y)$ satisfy Assumption \ref{ass:main-on-F}. Then for every accuracy $\eps$ and confidence parameter $\delta$ we have that
$$
\E ((\tilde{f}(X)-Y)^2 | (X_i,Y_i)_{i=1}^N) \leq \inf_{f \in F} \E(f(X)-Y)^2+\eps \ \ {\rm with \ probability \ } 1-\delta
$$
provided that $N \geq 2(N_1+N_2)$, where
\begin{align} \label{eq:sample-complexity-main}
N_1 = N_{\rm int}(T,\sqrt{\eps},c_1) + N_{\rm ext}(T,\sqrt{\eps},c_1) +c_2\left(\frac{L_T^2 \sigma^2}{\eps}+1\right) \log\left(\frac{64}{\delta}\right), \nonumber
\\
N_2 =  N_{\rm int}(\bar{T}_1,\sqrt{\eps},c_1) + N_{\rm ext}(\bar{T}_1,\sqrt{\eps},c_1) +c_2\left(\frac{L_T^2 \sigma^2}{\eps}+1\right) \log\left(\frac{64}{\delta}\right),
\end{align}
and $\sigma=\|f^*(X)-Y\|_{L_2}$.
\vskip0.3cm
The constants $c_1,c_2$ depend only on the uniform integrability function $\kappa$: if we set $\kappa_1=\kappa(1/10)$; $\xi_2 \sim {1}/{\kappa_1^2}$; and $\kappa_2=\kappa(\xi_2/4)$, then
$ c_1$ and $c_2$ depend only on $\kappa_1$ and $\kappa_2$.
\end{Theorem}
We describe the procedure in detail in Section \ref{sec:proc-in-detail}.
\end{framed}

Just like in \cite{LugMen16}, Theorem \ref{thm:main} has striking consequences: it implies that all the statistical content of a learning problem associated with the triplet $T=(F,X,Y)$ is actually coded in the `trivial' sample complexity $N_1+N_2$, which, by Claim \ref{claim:lower} and Claim \ref{claim:lower-2}, corresponds to the bare minimum one requires to overcome the trivial obstacles at a constant accuracy level.  And, once that minimal threshold is passed, the procedure requires only an additional sample whose cardinality is a lower bound on the sample complexity had the target been $Y=f_0(X)+W$ where $f_0 \in F$ and $W$ is a centred gaussian variable that is independent of $X$.

\begin{Remark}
Note that the procedure selects $\tilde{f} \in (F+F)/2$; hence, although in general the procedure is unrestricted, if $F$ happens to be convex then the procedure is actually proper. Moreover, in that case the estimate of Theorem \ref{thm:main} recovers the results from \cite{LugMen16}, though the procedure is completely different.
\end{Remark}

Of course, ${\bar F}_1$ is a random object, and to avoid having a data-dependent component in the sample complexity estimate one may simply take the largest sample complexity required for a set $H$ which satisfies that
\begin{equation} \label{eq:prop-F1}
f^* \in H  \subset \left(\frac{F+F}{2}\right),
\end{equation}
since $\bar{F}_1$ satisfies that condition. With that in mind, let us introduce the following notation.
\begin{Definition}
For a triplet $T=(F,X,Y)$, let ${\cal H}$ be the collection of all subsets $H$ of $(F+F)/2$ that
contain $f^*$. Set
$$
{\cal T}^\prime = \left\{ (H,X,Y) : H \in {\cal H}\right\}
$$
to be all the triplets associated with such classes $H$, the original distribution $X$ and the target $Y$.
\end{Definition}

Clearly, for Theorem \ref{thm:main} to hold it suffices that
$$
N_1, N_2 \geq \sup_{T^\prime \in {\cal T} } \left(N_{\rm int}(T^\prime,\sqrt{\eps},c_1) + N_{\rm ext}(T^\prime,\sqrt{\eps},c_1)\right) + c_2\left(\frac{L_T \sigma^2}{\eps}+1\right) \log\left(\frac{64}{\delta}\right),
$$
and often this upper bound is not much worst than \eqref{eq:sample-complexity-main}.

\begin{Remark}
Although Assumption \ref{ass:main-on-F} is rather natural, it does not cover one of the main families of problems encountered in learning theory: when the class consists of functions uniformly bounded by some constant $M$ and the target is also bounded by the same constant.

While Theorem \ref{thm:main} is not directly applicable to this \emph{bounded framework}, its proof actually is. In fact, because sums of iid bounded random variables exhibit a strong concentration phenomenon, the proof of a version of Theorem \ref{thm:main} that holds in the bounded framework is much simpler than in the general case we focus on here. Because our main interest is heavy-tailed problems, we only sketch the analogous result in the bounded framework in Appendix \ref{sec:bounded}.
\end{Remark}

To illustrate Theorem \ref{thm:main}, let us present the following classical example, which has been studied extensively in Statistics literature.

\subsection{Example -- finite dictionaries}
 One of the most important questions in modern high dimensional statistics has to do with prediction problems involving finite classes of functions or \emph{dictionaries}\footnote{In statistics literature, this is called  \emph{model-selection aggregation for a finite dictionary}. Aggregation problems of this type have been studied extensively over the years and we refer the reader  to \cite{Aud-unpub,MR2533466,MR2458184,LecMenAgg09,LRS-colt15,MenAgg16,MR3606751,Ts09} for more information on the subject.}. Because finite classes can never be convex, they fall out of the scope of \cite{LugMen16} and the resulting prediction problems call for a totally different approach.

For the sake of simplicity let illustrate the outcome of Theorem \ref{thm:main} by focusing on dictionaries in $\R^d$, i.e., for $t_1,...,t_m \in \R^d$, let $F=\{\inr{t_i,\cdot} : 1 \leq i \leq m\}$. Let $X$ be a centred random vector in $\R^d$ and as a working hypothesis (which can be relaxed further) assume that
\begin{description}
\item{$(1)$} There is a constant $L$ such that for every $t \in \R^d$, $\|\inr{X,t}\|_{L_4} \leq L\|\inr{X,t}\|_{L_2}$.
\item{$(2)$} The unknown target is of the form $Y=\inr{X,t_0}+W$, where $t_0 \in \R^d$ and $W$ is an unknown, mean-zero, square-integrable random variable that is independent of $X$.
\end{description}

\begin{Remark}
Clearly, the functions $\inr{X,t}$ can be heavy-tailed, as the norm equivalence only implies that $Pr(|\inr{X,t}| >u \|\inr{X,t}\|_{L_2})$ is slightly smaller than $L^4/u^4$. Also, since $W$ is just square integrable, $Y$ need not have any finite moment beyond the second one. This setup is totally out of reach for methods that exploit direct concentration arguments, and specifically, the results in \cite{LecMenAgg09,MR3606751}, which deal with dictionaries consisting of functions bounded by $1$ and targets that are bounded by $1$ are not applicable here.
\end{Remark}

For any such triplet $(F,X,Y)$ let $\inr{X,t^*}$ be the minimizer in $F$ of the squared risk functional and set $\sigma^2=\E (\inr{X,t^*}-Y)^2$. Applying Theorem \ref{thm:main} for a given accuracy $\eps$ and confidence parameter $\delta$, the procedure selects
$$
\tilde{t} \in \left\{\frac{t_i+t_j}{2} : 1 \leq i,j \leq m\right\};
$$
the constants $c_1$ and $c_2$ from Theorem \ref{thm:main} depend only on $L$, as does $L_T$. Hence, if
\begin{equation} \label{eq:finite-dictionary-accurate}
N \gtrsim_L \max_{T^\prime \in \tilde{T}^\prime} \left(N_{\rm int}(T^\prime,\sqrt{\eps},c_1) + N_{\rm ext}(T^\prime,\sqrt{\eps},c_1)\right)+\left(\frac{\sigma^2}{\eps}+1\right) \log\left(\frac{2}{\delta}\right),
\end{equation}
then
\begin{equation} \label{eq:finite-dictionary-intro-2}
\E \left((\inr{\tilde{t},X}-Y)^2 | (X_i,Y_i)_{i=1}^{N} \right) \leq \min_{1 \leq i \leq m} \E (\inr{X,t_i}-Y)^2 + \eps
\end{equation}
with probability at least $1-\delta$; the maximum in \eqref{eq:finite-dictionary-accurate} is with respect to all triplets
$$
{\cal T}^\prime = \left\{(H,X,Y) : H \subset (F+F)/2, \ \ {\rm and} \ \ \inr{X,t^*} \in H \right\}.
$$
As it happens, it is straightforward to obtain an upper estimate on \eqref{eq:finite-dictionary-accurate} that holds for any dictionary of cardinality $m$. For example, one may show that if $H$ is a dictionary consisting of $m$ points, $X$ is $L$-subgaussian\footnote{i.e., if in addition to being centred it satisfies that for any $t \in \R^d$ and $p \geq 2$, $\|\inr{X,t}\|_{L_p} \leq L\sqrt{p}\|\inr{X,t}\|_{L_2}$.}, and $Y=\inr{X,t_0}+W$ is as above, then
\begin{equation} \label{eq:worst-case-dictionary}
N_{\rm int}(T,\sqrt{\eps},c_1) + N_{\rm ext}(T,\sqrt{\eps},c_1) \leq c_3(L) \frac{\sigma^2}{\eps} \cdot \log m,
\end{equation}
where $c_3$ depends only on $L$. Hence, an upper estimate on \eqref{eq:finite-dictionary-accurate} that holds for any such triplet and in particular for any $T^\prime \in \cal{T}^\prime$ is that
\begin{equation} \label{eq:sample-finite-dictionary}
N_{\rm int}(T,\sqrt{\eps},c_1) + N_{\rm ext}(T,\sqrt{\eps},c_1) + \left(\frac{\sigma^2}{\eps}+1\right) \lesssim_L \left(\frac{\sigma^2}{\eps}+1\right) \log\left(\frac{2}{\delta}\right).
\end{equation}
It is well known that \eqref{eq:sample-finite-dictionary} is the best possible sample complexity estimate that holds for all possible dictionaries with $m$ points. Note, however, that \eqref{eq:sample-finite-dictionary} is attained after two significant steps that may come at a cost: first, in \eqref{eq:finite-dictionary-accurate} one replaces the triplet $(\bar{F}_1,X,Y)$ by the collection of triplets ${\cal T}^\prime$; and second, \eqref{eq:worst-case-dictionary} is a bound that holds for any dictionary of cardinality $m$, completely disregarding the geometry of the given class. Hence, \eqref{eq:sample-finite-dictionary} is a `worst-case' upper bound on the required sample complexity for the triplet $(F,X,Y)$. A better upper bound can be derived if one has more information on the structure of the dictionary, as its geometry is reflected in $N_{\rm int}$ and $N_{\rm ext}$.

We present a more detailed analysis of finite dictionaries, explain how their geometry affects the sample complexity and derive the worst-case bound \eqref{eq:worst-case-dictionary} in Appendix \ref{sec:dict}.

\vskip0.4cm

Despite being suboptimal, \eqref{eq:sample-finite-dictionary} is actually a considerable improvement on the current state-of-the-art in such problems, established in \cite{MenAgg16}. For example, let us compare the results from \cite{MenAgg16} to \eqref{eq:sample-finite-dictionary} in the case where $X$ is an $L$-subgaussian random vector and $Y =\inr{X,t_0}+W$ for some $t_0 \in \R^d$ and $W$ that is square-integrable and independent of $X$. 

\begin{Theorem} \cite{MenAgg16} \label{thm:dictionary-aggregation-old}
There is a procedure $\Psi$ for which the following holds. If we set $\hat{f}=\Psi\left((X_i,Y_i)_{i=1}^N\right)$, then
\begin{equation} \label{eq:example-error-rate}
\E ((\hat{f}(X)-Y)^2 | (X_i,Y_i)_{i=1}^N) \leq \inf_{f \in F} \E(f(X)-Y)^2+\eps \ \ {\rm with \ probability \ } 1-\delta
\end{equation}
provided that for some $q>2$
$$
N \geq c(L,q) \left(\frac{\|f^*(X)-Y\|_{L_q}}{\eps}+1\right) \log m \cdot {\rm poly}\left(\frac{1}{\delta}\right),
$$
and ${\rm poly}(1/\delta)$ scales like $(1/\delta)^{1/q}$ up to logarithmic factors.
\end{Theorem}

Both Theorem \ref{thm:dictionary-aggregation-old} and \eqref{eq:sample-finite-dictionary} deal with $X$ that is $L$-subgaussian and $Y=\inr{X,t_0}+W$, where $W$ can be heavy-tailed. Even if we take for granted that $\|f^*(X)-Y\|_{L_q} < \infty$ for some$q>2$ (which is not automatic; $W$ is assumed only to be square-integrable), it is clear that \eqref{eq:sample-finite-dictionary} is a much sharper estimate. Indeed, the clearest difference between Theorem \ref{thm:dictionary-aggregation-old} and \eqref{eq:sample-finite-dictionary} is the way the sample complexity scales with the confidence parameter $\delta$: the former is polynomial in $1/\delta$ and the latter is logarithmic in $1/\delta$.

The procedure $\Psi$ from Theorem \ref{thm:dictionary-aggregation-old} is suboptimal because it is based on \emph{Empirical Risk Minimization} (ERM), and ERM-based procedures perform poorly when faced with heavy-tailed data. ERM does reasonably well only when there are almost no outliers and the few existing outliers are not very far from the `bulk' of the data, but it does not cope well otherwise. Few and well-behaved outliers are to be expected only when the random variables involved have rapidly decaying tails (subgaussian) but when faced with data that is heavier tailed, like the $Y_i$'s in the example, ERM is bound to fail. We refer the reader to \cite{LugMen16} for a detailed discussion on ERM's sub-optimality, and turn now to describe a procedure that overcomes these issues. Like in \cite{LugMen16}, the procedure is based on a median-of-means tournament---though a very different one than the tournament used in \cite{LugMen16}.

\section{The procedure in detail} \label{sec:proc-in-detail}
The procedure we introduce is denoted by $\mP$ and consists of two components, $\mP_1$ and $\mP_2$.
\subsubsection*{$\mP_1$ -- estimating distances} \label{sec:mP_1}
The procedure $\mP_1$ receives as input a class of functions $H$ and a sample $(X_i)_{i=1}^N$. It has one tuning parameter, an integer $1 \leq \ell \leq N$.

\begin{framed}
\begin{Definition} \label{def:P-1}
For any pair of functions $h,f \in H$ and a sample $(X_i)_{i=1}^N$, set $v=(|f-h|(X_i))_{i=1}^N$ and let
$$
\mP_1(h,f)=v_\ell^*,
$$
where $(v_j^*)_{j=1}^N$ is the nonincreasing rearrangement of $(|v_j|)_{j=1}^N$.
\end{Definition}
\end{framed}

\subsubsection*{$\mP_2$ -- comparing statistical performance of functions}
The second component of the procedure receives as input a class $H$; a sample $(X_i,Y_i)_{i=N+1}^{2N}$; and all the outcomes $\mP_1(h,f)$ for $h,f \in H$, which were computed using the sample $(X_i)_{i=1}^N$.

$\mP_2$ has several tuning parameters, denoted by $\theta_1,\theta_2,\theta_3,\theta_{4}$ and it is also given the wanted accuracy and confidence parameters $\eps$ and $\delta$. Here and throughout this article, given the accuracy $\eps$ we set $r^2=c\eps$ for a constant $c$ that is specified in what follows. We also show that all the tuning parameters (including $c$) depend only on the uniform integrability function $\kappa(\xi)$ at two values: we set $\kappa_1=\max\{\kappa(1/10),1\}$, and for $\xi_2=1/\kappa_1^2$ we set $\kappa_2=\kappa(\xi_2^2/4)$; the tuning parameters depend only on $\kappa_1$ and $\kappa_2$.

\vskip0.4cm

To define $\mP_2$, Let
$$
n=\theta_1\log\left(\frac{64}{\delta}\right) \ \ \ \ {\rm and \ set} \ \ \ \ m=\frac{N}{n}.
$$

We split $\{1,...,N\}$ to $n$ coordinate blocks $I_1,...,I_n$ which, without loss of generality are assumed to be of equal size, denoted by $m$.


For $h,f \in H$ and $1 \leq j \leq n$ let
$$
\B_{h,f}(j)=\frac{1}{m}\sum_{i \in I_j} (h(X_i)-Y_i)^2 - \frac{1}{m}\sum_{i \in I_j} (f(X_i)-Y_i)^2.
$$
\begin{framed}
\begin{Definition} \label{def:win}
Set $f \succ h$ if, for more than $n/2$ of the coordinate blocks $I_j$, one has
\begin{equation}
\begin{cases}
\B_{h,f}(j) \geq -\theta_2 r^2 & \mbox{when} \ \ \ \mP_1(h,f) \leq \theta_4 r, \ \ \ {\rm or}
\\
\B_{h,f}(j) \geq -\theta_3 \mP_1^2(h,f) & \mbox{when} \ \ \ \mP_1(h,f)>\theta_4 r.
\end{cases}
\end{equation}
Let
\begin{equation} \label{eq:P-2-def}
\mP_2(H)=\left\{f \in H : \ f \succ h \ {\rm for \ every } \ h \in H\right\}.
\end{equation}
\end{Definition}
\end{framed}

It is a little easier to follow the meaning of Definition \ref{def:win} if one thinks of $\mP_2$ as a tournament procedure, and Definition \ref{def:win} as representing the outcome of a `home-and-way' type match between any two elements in $H$: the function $f$ wins its home match against $h$ if $f \succ h$. Therefore, $\mP_2(H)$ consists of all the functions in $H$ that have won all their home matches in the tournament. Note that it is possible to have both $h \succ f$ and $f \succ h$.

\vskip0.3cm
Therefore, the complete procedure is as follows:
\begin{framed}
Given the triplet $(F,X,Y)$ that satisfies Assumption \ref{ass:main-on-F} and the sample $(X_i,Y_i)_{i=1}^{4N}$,
\begin{description}
\item{(1)} Run $\mP_1$ on $F$ using $(X_i)_{i=1}^N$ as its input.
\item{(2)} Using the outcome of $(1)$ and $(X_i,Y_i)_{i=N+1}^{2N}$ as input, run $\mP_2$ on $F$. Set $F_1=\mP_2(F)$ (and observe that $F_1$ is a subset of $F$).
\item{(3)} Let $\bar{F}_1=\{(f+h)/2 : f,h \in F_1\}$ and repeat $(1)$ and $(2)$ for the class $\bar{F}_1$ using the samples $(X_i)_{i=2N+1}^{3N}$ and $(X_i,Y_i)_{i=3N+1}^{4N}$, respectively.
\item{(4)} Let $F_2=\mP_2(\bar{F}_1)$ and select any function $\tilde{f} \in F_2$.
\end{description}
Moreover, recalling that
$$
\kappa_1=\kappa(1/10); \ \ \xi_2 \sim \frac{1}{\kappa_1^2} \ \ {\rm and} \ \ \kappa_2=\kappa(\xi_2/4),
$$
we have that $r \sim \eps/\kappa_1^2$ and the tuning parameters the procedure uses are
$$
\ell=\frac{N}{5 \kappa_1^2}; \ \ \theta_1 \sim \frac{\xi_2^2}{\kappa_2^2}; \ \ \theta_2 \sim \kappa_1^2; \ \ \theta_3 \sim \frac{1}{\kappa_1^2}; \ \ \theta_4 \sim \kappa_1.
$$
\end{framed}


\begin{Remark}
Estimating distances between class members has nothing to do with the unknown target $Y$ and does not require the ``labels" $Y_i$. Thus, $\mP_1$ may be considered as a pre-processing step and can then be used for any target $Y$---simply by running $\mP_1$ for the class $(F+F)/2=\{(f+h)/2 : f,h \in F\}$. In fact, there is nothing special about $\mP_1$; it may be replaced by any data-dependent procedure that satisfies $\|h-h^*\|_{L_2} \lesssim \mP_1(h,h^*) \lesssim C\|h-h^*\|_{L_2}$ as long as $\|h-h^*\|_{L_2} \geq c^\prime r$ (recall that $r^2 \sim \eps$, the required accuracy level). This fact will be of use when we explore the bounded framework, in Appendix \ref{sec:bounded}.

Finally, there are some situations in which $\mP_1$ is not needed at all. For example, if $X$ is a random vector in $\R^d$ with independent, mean-zero, variance $1$ random variables as coordinates then its covariance structure coincides with the standard Euclidean structure in $\R^d$. Thus, for any $t_1,t_2 \in \R^d$, $\|\inr{X,t_1-t_2}\|_{L_2}=\|t_1-t_2\|_2$ and there is no need to estimate the $L_2$ distances between linear functionals $\inr{t,\cdot}$.
\end{Remark}

\section{Proof of Theorem \ref{thm:main}} \label{sec:proofs}
The proof of Theorem \ref{thm:main} is rather involved and technical, and requires some unpleasant `constant chasing'; unfortunately, that cannot be helped. We begin its presentation with a short road-map, outlining the argument.

The first component in the proof is a reduction step: identifying a sufficient (random) condition, which, once verified, implies that the procedure performs as expected. This reduction step is presented in Section \ref{sec:interlude} and Section \ref{sec:det-to-rand}.

The study of the random condition is the heart of the matter. To prove its validity with the desired confidence, one has to show that the quadratic and multiples components of the excess risk functional are `regular' in an appropriate sense once the sample size is large enough. Proving that is the topic of Section \ref{sec:proof-rand}.

\vskip0.4cm

To ease some of the technical difficulty in the proofs of the random components from Section \ref{sec:proof-rand}, it is helpful to keep in mind the following facts:
\begin{description}
\item{$\bullet$} All the constants appearing in the proof are derived from the uniform integrability function $\kappa(\xi)$ at two different, well specified levels. Although we keep track of those constants, one should realize that since $\kappa(\xi)$ is known, they are just fixed numbers, and of limited significance to the understanding of what is going on.

\item{$\bullet$} The number of coordinate blocks used in the tournament is $n \sim \log(64/\delta)$. The motivation behind this choice is simple: if a certain property holds for a single function on an individual block with constant probability, then by the independence of the blocks the probability that the property is satisfied by a majority of the blocks is exponential in $n$. With our choice of $n$, the resulting confidence is $1-2\exp(cn)=1-\delta/C$, which is precisely what we are looking for.

\item{$\bullet$} The choice of the sample size $N$ is made to ensure that one has `enough randomness', leading to a regular behaviour of the random variables involved in the proof. We establish a quantitative estimate on the sample size that is needed for that regularity, but it is instructive to note as the proof progresses that the wanted control becomes more likely as $N$ increases.
\end{description}

\subsection{A deterministic interlude} \label{sec:interlude}
There is a feature that plays an important role in most unrestricted procedures: if one can find two almost minimizers of the risk that are far apart, their midpoint is much closer to $Y$ than $f^*$ is. Each procedure looks for such functions and exploits their existence in a different way, but up to this point, all the methods that have been used to that end were based on empirical minimization. This deterministic interlude is a step towards an alternative path: finding, without resorting to ERM, a subset of the given class that consists of functions that are either very close to $f^*$, or that their average with $f^*$ is significantly closer to $Y$ than $f^*$ is.

\vskip0.3cm

Let $(F,X,Y)$ be a triplet, fix $r>0$ and recall that $\sigma=\|f^*(X)-Y\|_{L_2}$. Let $\HH$ be the hyperplane supporting the ball $\{h  : \|h(X)-Y\|_{L_2} \leq \sigma\}$ at $f^*$. Observe that if $F$ happens to be convex then $F \subset \HH^+$---the `positive side' of $\HH$, defined by the condition $\E (h(X)-f^*(X)) \cdot (f^*(X)-Y) \geq 0$. Indeed, this follows from the characterization of the nearest point map onto a closed, convex subset of a Hilbert space.

Of course, $F$ need not be convex. Therefore, as a preliminary goal one would like to identify a subset of $F$, containing $f^*$ and possibly other functions as well, as long as they satisfy the following:
\begin{description}
\item{$(1)$} If $f \in F \cap \HH^+$ then $f$ is an almost minimizer of the risk, in the sense that $\|f(X)-Y\|_{L_2}^2 \leq \|f^*(X)-Y\|_{L_2}^2 + r^2$.
\item{$(2)$} If $f \in F \cap \HH^-$ and $h = (f+f^*)/2$ then $\|h(X)-Y\|_{L_2}$ is significantly smaller than $\|f^*(X)-Y\|_{L_2}$.
\end{description}


We call such a subset an \emph{essential part} of the class, though it depends on the entire triplet:
\begin{Definition} \label{def:essential}
Let $(H,X,Y)$ be a triplet. For $r>0$ and $0 < \rho < 1$, a subset $H^\prime \subset H$ is $(\rho,r)$-essential if $h^* \in H^\prime$ and for every $h \in H^\prime$,
\begin{equation} \label{eq:essential}
\|h(X)-Y\|_{L_2}^2 \leq \|h^*(X)-Y\|_{L_2}^2 + \rho \|h-h^*\|_{L_2}^2 + r^2.
\end{equation}
\end{Definition}
Observe that \eqref{eq:essential} amounts to
\begin{equation} \label{eq:equiv-essential}
(1-\rho) \|h-h^*\|_{L_2}^2 \leq -2 \E (h-h^*)(X) \cdot (h^*(X)-Y)  +r^2.
\end{equation}
Therefore, if $0<\rho <1$, functions in the class that belong to $\HH^+$ can satisfy \eqref{eq:essential} only if $\|h-h^*\|_{L_2}^2 \leq r^2/(1-\rho)$; moreover, the condition becomes harder to fulfill the further $h \in \HH^+$ is from the hyperplane $\HH$. On the other hand, functions in the class that belong to  $\HH^-$ satisfy \eqref{eq:essential} only if there is a balance between their distance to $h^*$ and the `direction' of the interval connecting them to $h^*$. With the right balance one may show that the mid-point $(h+h^*)/2$ is much closer to $Y$ than $h^*$ is.

A crucial observation is that identifying an essential part of a class is all that is needed if one is interested in finding a function with a small excess risk. Indeed, assume one has a (deterministic) procedure, denoted by $\mP_0$,  that receives as input a triplet $(H,X,Y)$ and values $r>0$ and $0 < \rho <1/18$, and returns a set $\mP_0(H)$ that is a $(\rho,r)$-essential subset of $H$ (note that despite the minor abuse of notation, $\mP_0$ depends on the whole triplet and not just on the underlying class).

\begin{Theorem} \label{thm:essential-to-approximation}
Let $(F,X,Y)$ be a triplet. Set $F_1=\mP_0(F)$, let $\bar{F}_1=(F_1+F_1)/2$ and for the triplet $(\bar{F}_1,X,Y)$, let $F_2=\mP_0\left(\bar{F}_1\right)$.

If $\rho \leq 1/18$ then every $f \in F_2$ satisfies
\begin{equation} \label{eq:in-thm-ess-to-approx}
\|f(X)-Y\|_{L_2}^2 \leq \inf_{f \in F} \|f(X)-Y\|_{L_2}^2 + \frac{3r^2}{2}.
\end{equation}
\end{Theorem}

\proof
Consider a generic triplet $(H,X,Y)$ and set $H^\prime = \mP_0(H)$. The heart of the proof is the following observation: let $0<\rho \leq 1/2$;
\begin{description}
\item{$\bullet$} if ${\rm diam}(H^\prime,L_2) \leq 2r$ then for every $h \in H^\prime$,
$$
\|h(X)-Y\|_{L_2}^2 \leq \|h^*(X)-Y\|_{L_2}^2 + (4\rho +1)r^2;
$$,
\item{$\bullet$} otherwise, if ${\rm diam}(H^\prime,L_2) \geq 2r$, there is some $u \in (H^\prime+h^*)/2 \subset (H^\prime+H^\prime)/2$ such that
$$
\|u(X)-Y\|_{L_2}^2 - \|h^*(X)-Y\|_{L_2}^2 \leq -\frac{1}{16}(1-2\rho){\rm diam}^2(H^\prime,L_2) + \frac{r^2}{2}.
$$
\end{description}

Indeed, on the one hand, if ${\rm diam}(H^\prime,L_2) \leq 2r$, then since $H^\prime$ is $(\rho,r)$-essential,
$$
\|h(X)-Y\|_{L_2}^2 \leq \|h^*(X)-Y\|_{L_2}^2 + (4\rho +1)r^2.
$$
On the other hand, if ${\rm diam}(H^\prime,L_2) \geq 2r$, there is some $h \in H^\prime$ that satisfies $\|h-h^*\|_{L_2} \geq {\rm diam}(H^\prime,L_2)/2 \geq r$. Set $u=(h+h^*)/2$ and note that
\begin{align*}
& \|u(X)-Y\|_{L_2}^2 =  \frac{1}{4}\|h(X)-Y\|_{L_2}^2 + \frac{1}{4}\|h^*(X)-Y\|_{L_2}^2 + \frac{1}{2}\E(h(X)-Y) \cdot (h^*(X)-Y)
\\
= & \frac{1}{4}\|h(X)-Y\|_{L_2}^2 + \frac{1}{4}\|h^*(X)-Y\|_{L_2}^2 + \frac{1}{2}\E(h(X)-h^*(X)) \cdot (h^*(X)-Y) + \frac{1}{2}\|h^*(X)-Y\|_{L_2}^2.
\end{align*}
Therefore,
\begin{equation} \label{eq:0}
\|u(X)-Y\|_{L_2}^2 - \|h^*(X)-Y\|_{L_2}^2  = \frac{1}{4}\|h(X)-Y\|_{L_2}^2 - \frac{1}{4}\|h^*(X)-Y\|_{L_2}^2 + \frac{1}{2}\E(h(X)-h^*(X)) \cdot (h^*(X)-Y).
\end{equation}
Also,
\begin{equation} \label{eq:1}
\|h(X)-Y\|_{L_2}^2 - \|h^*(X)-Y\|_{L_2}^2 = \|h-h^*\|_{L_2}^2 + 2 \E(h(X)-h^*(X)) \cdot (h^*(X)-Y);
\end{equation}
because $H^\prime$ is $(\rho,r)$-essential it follows that
\begin{equation*}
\|h(X)-Y\|_{L_2}^2 - \|h^*(X)-Y\|_{L_2}^2 \leq \rho \|h-h^*\|_{L_2}^2 + r^2,
\end{equation*}
and in particular,
\begin{equation} \label{eq:2}
\E(h(X)-h^*(X)) \cdot (h^*(X)-Y) \leq -\frac{1}{2}(1-\rho)\|h-h^*\|_{L_2}^2 + \frac{r^2}{2}.
\end{equation}
Combining \eqref{eq:0}, \eqref{eq:1} and \eqref{eq:2}, and since $0<\rho \leq 1/18$, it follows that
\begin{align*}
\|u(X)-Y\|_{L_2}^2 - \|h^*(X)-Y\|_{L_2}^2 \leq & \frac{1}{4}\|h-h^*\|_{L_2}^2 +  \E(h(X)-h^*(X)) \cdot (h^*(X)-Y)
\\
\leq & \left(-\frac{1}{4}+\frac{\rho}{2}\right)\|h-h^*\|_{L_2}^2 + \frac{r^2}{2}
\\
\leq & \left(-\frac{1}{4}+\frac{\rho}{2}\right)\frac{{\rm diam}^2(H^\prime,L_2)}{4} + \frac{r^2}{2}.
\end{align*}

Now, given the class $F$, let $F_1=\mP_0(F)$, set $\bar{F}_1=(F_1+F_1)/2$ and consider $F_2=\mP_0(\bar{F}_1)$. In the first alternative, ${\rm diam}(F_1,L_2) \leq 2r$; for every $f \in F_1$
\begin{equation} \label{eq:3}
\|f(X)-Y\|_{L_2}^2 \leq \|f^*(X)-Y\|_{L_2}^2 + (4\rho +1)r^2;
\end{equation}
and by the convexity of $\| \ \|_{L_2}^2$, \eqref{eq:3} holds for every function in $\bar{F}_1$. \eqref{eq:in-thm-ess-to-approx} follows because $F_2 \subset \bar{F}_1$.

Otherwise, if ${\rm diam}(F_1,L_2) \geq 2r$, let $h^*={\rm argmin}_{f \in \bar{F}_1} \|f(X)-Y\|_{L_2}$ and recall that $f^* \in F_1$. Applying the second alternative,
\begin{equation} \label{eq:improved-approximation}
\|h^*(X)-Y\|_{L_2}^2 - \|f^*(X)-Y\|_{L_2}^2 \leq -\frac{1}{16}(1-2\rho){\rm diam}^2(F_1,L_2) + \frac{r^2}{2}.
\end{equation}
Observe that ${\rm diam}(F_1,L_2)={\rm diam}(\bar{F}_1,L_2)$. Since $F_2$ is a $(\rho,r)$-essential subset of $\bar{F}_1$ and invoking \eqref{eq:improved-approximation}, it follows that for every $f \in F_2$,

\begin{align*}
\|f(X)-Y\|_{L_2}^2 \leq & \|h^*(X)-Y\|_{L_2}^2 + \rho \|f-h^*\|_{L_2}^2 + r^2
\\
\leq & \|f^*(X)-Y\|_{L_2}^2 -\frac{1}{16}(1-2\rho){\rm diam}^2(\bar{F}_1,L_2)+\rho \|f-h^*\|_{L_2}^2 + \frac{3r^2}{2}
\\
\leq & \|f^*(X)-Y\|_{L_2}^2 -\frac{1}{16}(1-18\rho){\rm diam}^2(\bar{F}_1,L_2) + \frac{3r^2}{2}
\\
\leq & \|f^*(X)-Y\|_{L_2}^2 + \frac{3r^2}{2},
\end{align*}
provided that $\rho \leq 1/18$.
\endproof

Thanks to Theorem \ref{thm:essential-to-approximation}, the proof of Theorem \ref{thm:main} is reduced to showing that our data-dependent procedure $\mP$, when given an arbitrary triplet $(H,X,Y)$, generates an $(1/20,r)$-essential subset of $H$ with probability at least $1-\delta/2$ where $r^2 \sim \eps$. Indeed, first set $H_1=F$ and ensure that with the required probability, for $(X_i,Y_i)_{i=N+1}^{2N}$, the output $F_1 = \mP_2(F)$ is a $(1/20,r)$-essential subset of $F$, and then, for $H_2=\{(f+h)/2 : f,h \in F_1\}$, ensure that $F_2=\mP_2(H_2)$ is a $(1/20,r)$-essential subset of $H_2$. Theorem \ref{thm:essential-to-approximation} implies that any $\tilde{f} \in F_2$ satisfies that
\begin{equation} \label{eq:from-essential-to-error}
\|\tilde{f}(X)-Y\|_{L_2}^2 \leq \|f^*(X)-Y\|_{L_2}^2 + \frac{3r^2}{2}.
\end{equation}

Therefore, all that is needed to complete the proof of Theorem \ref{thm:main} is the following claim:
\begin{Claim} \label{claim:reduction}
For any triplet $(H,X,Y)$ that satisfies Assumption \ref{ass:main-on-F}, and setting $r^2 \sim \eps$, there is an event of probability at least $1-\delta/2$, for which
\begin{description}
\item{$\bullet$} the true minimizer $h^*$ wins all of it `home games' in the sense of Definition \ref{def:win}, and
\item{$\bullet$} if $h$ wins all of its home games then it satisfies that
$$
\|h(X)-Y\|_{L_2}^2 \leq \|h^*(X)-Y\|_{L_2}^2 + \frac{1}{20} \|h-h^*\|_{L_2}^2 + r^2.
$$
\end{description}
\end{Claim}
Indeed, Claim \ref{claim:reduction} implies that the set of `winners' in our tournament is the wanted $(1/20,r)$-essential subset of $H$.

Of course, a central part of the proof of Claim \ref{claim:reduction} is to identify the tuning parameters $\ell, \theta_1,...,\theta_4$ needed in the definition of the components $\mP_1$ and $\mP_2$.

\subsection{From deterministic to random} \label{sec:det-to-rand}

The proof of Claim \ref{claim:reduction} calls for two additional steps. The first, presented in Section \ref{sec:det-to-rand}, is another reduction step: identifying sufficient estimates on certain random processes, that once verified, imply that the combination of $\mP_1$ and $\mP_2$ generates a $(1/20,r)$-essential subset of $H$ for an arbitrary triplet $(H,X,Y)$. Then, in Section \ref{sec:proof-rand}, we confirm that the required probabilistic estimates are indeed true.

In addition to establishing control on certain Rademacher averages (appearing in the definition of $N_{\rm int}$ and $N_{\rm ext}$), we also require information on the packing numbers of localizations of $H$.

\begin{Definition} \label{def:pack}
Given a set $H \subset L_2(\mu)$ and $\eps>0$, let ${\cal M}(H,\eps D)$ be the
$\eps$-packing number of $H$. Thus, ${\cal M}(H,\eps D)$
is the maximal cardinality of a subset $\{h_1,...,h_m\} \subset H$ such that $\|h_i-h_j\|_{L_2} \geq \eps$ for every $i \not = j$.
\end{Definition}

Along the proof of Claim \ref{claim:reduction}, we collect conditions on the sample size $N$, ensuring that
\begin{equation} \label{eq:packing-condition}
\log {\cal M}(H_{h^*,r},\gamma_1 rD) \leq \gamma_2 N;
\end{equation}

\begin{equation} \label{eq:Rad-Q-condition}
\E \sup_{u \in H_{h^*,r}} \left|\frac{1}{\sqrt{N}} \sum_{i=1}^N \eps_i u(X_i) \right| \leq \gamma_3 \sqrt{N} r;
\end{equation}
and
\begin{equation} \label{eq:Rad-M-condition}
\E \sup_{u \in H_{h^*,r}} \left|\frac{1}{\sqrt{N}} \sum_{i=1}^N \eps_i (h^*(X_i)-Y_i) u(X_i) \right| \leq \gamma_4 \sqrt{N} r^2
\end{equation}
for various constants $\gamma_1,...,\gamma_4$.

Intuitively, \eqref{eq:packing-condition}, \eqref{eq:Rad-Q-condition} and \eqref{eq:Rad-M-condition} become `easier' the larger $N$ is. In fact, it is standard to show that if either \eqref{eq:packing-condition}, \eqref{eq:Rad-Q-condition} or \eqref{eq:Rad-M-condition} holds for an integer $N_0$, then it necessarily holds for every $N \geq N_0$ --- with the constants $\gamma_3$ and $\gamma_4$ replaced by $2\gamma_3$ and $2\gamma_4$ in \eqref{eq:Rad-Q-condition} and \eqref{eq:Rad-M-condition} respectively. Let us illustrate this for \eqref{eq:Rad-Q-condition}: if $N=2^k$ and
$$
\E \sup_{u \in H_{h^*,r}} \left|\sum_{i=1}^{2^k} \eps_i u(X_i) \right| \leq \gamma_3 2^{k} r
$$
then by the triangle inequality a similar estimate holds for $N=2^{k+1}$. And if $2^k \leq N \leq 2^{k+1}$ then set $a_i=1$ for $1 \leq i \leq N$ and $a_i = 0$ otherwise. By a contraction argument for Bernoulli processes (see, e.g. \cite{LeTa91}),
\begin{align*}
& \E \sup_{u \in H_{h^*,r}} \left|\sum_{i=1}^{N} \eps_i u(X_i) \right| = \E \sup_{u \in H_{h^*,r}} \left|\sum_{i=1}^{2^{k+1}} \eps_i a_i u(X_i) \right|
\leq \E \sup_{u \in H_{h^*,r}} \left|\sum_{i=1}^{2^{k+1}} \eps_i u(X_i) \right|
\\
\leq & \gamma_3 2^{k+1} r \leq 2\gamma_3 Nr.
\end{align*}

This observation gives one the freedom to increase $N$ without worrying that \eqref{eq:packing-condition}, \eqref{eq:Rad-Q-condition} or \eqref{eq:Rad-M-condition} become invalid once they have been established for some integer $N_0$.

A similar regularity holds with respect to $r$ and to $\gamma_1,...,\gamma_4$. The fact that $H_{h^*,r}$ is star-shaped around $0$ ensures that once a condition is satisfied by $r$, it is automatically satisfied by any $r^\prime > r$. For the same reason, the  conditions are `monotone' in the parameters $\gamma_1,...,\gamma_4$, allowing one to use the smallest, most restrictive constants one has collected along the way. The proof of this type of regularity can be found in \cite{LugMen16} and will not be presented here.

\vskip0.4cm
Unlike \eqref{eq:Rad-Q-condition} and \eqref{eq:Rad-M-condition},  \eqref{eq:packing-condition} has not appeared in the sample complexity estimate stated in Theorem \ref{thm:main}. We will show in Theorem \ref{thm:Bernoulli-implies-entropy} that \eqref{eq:packing-condition} is actually implied by \eqref{eq:Rad-Q-condition} for a well chosen $\gamma_3$ once the class satisfies Assumption \ref{ass:main-on-F}.

Let us formulate a sufficient condition that ensures that our data-dependent procedure produces a $(\rho,r)$-essential subset of $H$ for an arbitrary triplet $(H,X,Y)$.

Recall the sample size $N$ satisfies that
\begin{equation} \label{eq:cond-on-N}
N \geq \theta_0 \max\left\{\frac{L_T^2\sigma^2}{r^2},1 \right\} \log \left(\frac{64}{\delta}\right),
\end{equation}
where, as always, $r^2 \sim \eps$ --- the wanted accuracy, and $\theta_0$ is a constant that will be specified in what follows. Let $I_1,...,I_n$ is a partition of $\{1,...,N\}$ to $n$ coordinate blocks, where
\begin{equation} \label{eq:cond-on-n}
n=\theta_1 \log\left(\frac{64}{\delta}\right)
\end{equation}
for a well-chosen constant $\theta_1$. Thus, each one of the blocks is of cardinality $m=N/n$, and with our choice of $N$,
\begin{equation} \label{eq:condition-on-m}
\frac{1}{m} =\frac{n}{N} \leq \frac{\theta_1}{\theta_0} \min\left\{\frac{r^2}{L_T^2 \sigma^2},1\right\}.
\end{equation}

For every $h,f \in H$, set
\begin{align*}
\M_{h,f}(j) = & \frac{2}{m} \sum_{i \in I_j} (h(X_i)-f(X_i))(f(X_i)-Y_i), \ \ \ \
\Q_{h,f}(j) = \frac{1}{m} \sum_{i \in I_j} (h-f)^2(X_i),
\\
\B_{h,f}(j) = & \frac{1}{m} \sum_{i \in I_j} (h(X_i)-Y_i)^2 - \frac{1}{m} \sum_{i \in I_j} (f(X_i)-Y_i)^2 = \Q_{h,f}(j)+\M_{h,f}(j).
\end{align*}
In what follows we write $\E \M_{h,f}$ instead of $\E \M_{h,f}(j)$ as all these expectations do not change with $j$.

\vskip0.3cm

Recall that $\mP_1$ is a procedure that, given a sample $(X_i)_{i=1}^N$ and $h,f \in H$, returns the value $\mP_1(h,f)=\mP_1(f,h)$. Intuitively, $\mP_1$ serves as an estimator of $L_2$ distances between class members, and to give a quantitative meaning to this intuition, fix $\alpha<1$ and $\beta>1$; let ${\cal A}^\prime$ be an event for which the following holds:
\begin{framed}
For any $h \in H$:
\begin{description}
\item{$(1)$} If $\mP_1(h,h^*) \geq \beta r$ then
$$
\beta^{-1}\mP_1(h,h^*) \leq \|h-h^*\|_{L_2} \leq \alpha^{-1}\mP_1(h,h^*),
$$
and if $\mP_1(h,h^*) < \beta r$ then $\|h-h^*\|_{L_2} \leq (\beta/\alpha)r$.
\end{description}
\end{framed}
In other words, on ${\cal A}^\prime$, if $\mP_1(h,h^*)$ is large enough, then $\mP_1$ is a two-sided isomorphic estimate of $\|h-h^*\|_{L_2}$, and otherwise, $h$ and $h^*$ are relatively close.


\vskip0.3cm

Turning to $\mP_2$, let $\gamma>0$ and $0<\nu<1$ and set ${\cal A}^{\prime \prime}$ be the event for which
\begin{framed}
for every $h \in H$,
\begin{description}
\item{$(2)$} If $\|h-h^*\|_{L_2} \geq r$ then on more than $n/2$ of the coordinate blocks $I_j$,
$$
\Q_{h,h^*}(j) \geq (1-\nu)\|h-h^*\|_{L_2}^2 \ \ {\rm and} \ \ \M_{h,h^*} - \E \M_{h,h^*} \geq -\nu \|h-h^*\|_{L_2}^2.
$$
\item{$(3)$} If $\|h-h^*\|_{L_2} \leq (\beta/\alpha)r$ then on more than $n/2$ of the coordinate blocks
$$
|\M_{h,h^*}(j)-\E \M_{h,h^*} | \leq \gamma r^2.
$$
\end{description}
\end{framed}

With those conditions set in place, let us select the tuning parameters $\theta_2$, $\theta_3$ and $\theta_4$ accordingly, stating how they depend on $\alpha$, $\beta$, $\gamma$ and $\nu$.  Set
$$
\theta_2= \frac{\beta^2}{\alpha^2} + \gamma; \ \ \ \theta_3 = \frac{2\nu}{\alpha^2}; \ \ \ \theta_4 =\beta.
$$
Thus, $\mP_2$ receives as input the values $\mP_1(f,h)$ for any $f,h \in H$, obtained using the sample $(X_i)_{i=1}^N$. And, given an independent sample $(X_i,Y_i)_{i=N+1}^{2N}$,
\begin{framed}
$f \succ h$ if for more than $n/2$ of the coordinate blocks $I_j$,
\begin{equation}
\begin{cases}
\B_{h,f}(j) \geq -((\beta/\alpha)^2+\gamma)r^2 & \mbox{when} \ \ \ \mP_1(h,f) < \beta r, \ \ and
\\
\B_{h,f}(j) \geq -(2\nu/\alpha^2) \mP_1^2(h,f) & \mbox{when } \ \ \ \mP_1(h,f) \geq \beta r.
\end{cases}
\end{equation}
Recall that the output of the procedure is
\begin{equation}
\mP_2(H)=\left\{f \in H : \ f \succ h \ {\rm for \ every } \ h \in H\right\}.
\end{equation}
\end{framed}

The main observation is that on the event ${\cal A}^\prime \cap  {\cal A}^{\prime \prime}$, $\mP$ generates an essential subset of $H$. 
\begin{Theorem} \label{thm:observation}
Set ${\cal A}={\cal A}^\prime \cap {\cal A}^{\prime \prime}$. For every sample $(X_i,Y_i)_{i=1}^{2N} \in {\cal A}$,
\begin{description}
\item{$\bullet$} $h^* \in \mP_2(H)$, and,
\item{$\bullet$} if $h \in \mP_2(H)$ then
\begin{equation} \label{eq:observation}
\|h(X)-Y\|_{L_2}^2 \leq \|h^*(X)-Y\|_{L_2}^2 + 2\nu \left(1+\frac{\beta^2}{\alpha^2}\right) \|h-h^*\|_{L_2}^2 + 2\left(\gamma+\frac{\beta^2}{\alpha^2}\right)r^2.
\end{equation}
\end{description}
In particular, on the event ${\cal A}$, $\mP_2(H)$ is a $(\rho,r^\prime)$-essential subset of $H$ for
\begin{equation} \label{eq:in-thm-cond-essential}
\rho = 2\nu \left(1+\frac{\beta^2}{\alpha^2}\right) \ \ \ {\rm and} \ \ \ r^\prime = \sqrt{2}\left(\gamma+\frac{\beta^2}{\alpha^2}\right)^{1/2}r.
\end{equation}
\end{Theorem}

Once Theorem \ref{thm:observation} is established, the path towards a proof of Claim \ref{claim:reduction} is clear: one has to show that there are parameters $\alpha,\beta, \gamma$ and $\nu$ that satisfy the conditions of the Theorem \ref{thm:observation} for sample size $N$ as stated in Theorem \ref{thm:main}; that $Pr({\cal A}) \geq 1-\delta/2$; and that $\nu$ can be taken small enough to ensure that $\rho = 2\nu (1+\beta^2/\alpha^2) \leq 1/20$. Also, one has to specify the two missing tuning parameters: $\ell$, which appears in the definition of $\mP_1$ and $\theta_1$ which appears in the choice of the number of coordinate blocks $n$.

\vskip0.3cm
\noindent{\bf Proof of Theorem \ref{thm:observation}.}
Note that for $h \in H$,
\begin{align*}
\B_{h,h^*}(j) = & \frac{1}{m}\sum_{i \in I_j} \left(h(X_i)-Y_i\right)^2 - \frac{1}{m}\sum_{i \in I_j} \left(h^*(X_i)-Y_i\right)^2
\\
= & \Q_{h,h^*}(j) + \left(\M_{h,h^*}(j) - \E\M_{h,h^*}(j)\right) + 2\E(h-h^*)(X)(h(X)-Y),
\end{align*}
and setting the excess risk function ${\cal L}_h(X,Y) = (h(X)-Y)^2 - (h^*(X)-Y)^2$, it follows that
\begin{equation} \label{eq:B-in-proof}
\B_{h,h^*}(j) = \E{\cal L}_h +  \left(\Q_{h,h^*}(j)-\|h-h^*\|_{L_2}^2\right) + \left(\M_{h,h^*}(j) - \E\M_{h,h^*}(j)\right).
\end{equation}
Assume first that $\mP_1(h,h^*) \geq \beta r$. By Condition $(1)$, $\|h-h^*\|_{L_2} \geq r$, and by Condition $(2)$,
$$
\Q_{h,h^*}(j) \geq (1-\nu)\|h-h^*\|_{L_2}^2 \ \ {\rm and} \ \ \M_{h,h^*}(j) - \E\M_{h,h^*}(j) \geq -\nu \|h-h^*\|_{L_2}^2
$$
on more than $n/2$ of the coordinate blocks. For such a coordinate block $j$,
$$
\B_{h,h^*}(j) \geq \E{\cal L}_h -2\nu \|h-h^*\|_{L_2}^2 \geq -2\nu \|h-h^*\|_{L_2}^2 \geq -\frac{2\nu}{\alpha^2}\mP_1^2(h,h^*),
$$
which is evident because $\E{\cal L}_h \geq 0$ and by Condition $(1)$, $\|h-h^*\|_{L_2} \leq \alpha^{-1}\mP_1(h,h^*)$. Therefore, in this case $h^* \succ h$.

If, on the other hand, $\mP_1(h,h^*) < \beta r$ then by Condition $(1)$, $\|h-h^*\|_{L_2} \leq (\beta/\alpha)r$. Clearly, both $\E {\cal L}_h$ and $\Q_{h,h^*}(j)$ are nonnegative and by condition $(3)$, $|\M_{h,f}(j)-\E \M_{h,f} | \leq \gamma r^2$ on more than $n/2$ of the coordinate blocks. Hence, on these blocks,
$$
\B_{h,h^*}(j) \geq -\|h-h^*\|_{L_2}^2 - |\M_{h,h^*}(j) - \E\M_{h,h^*}(j)| \geq -\left(\frac{\beta^2}{\alpha^2} + \gamma\right)r^2,
$$
and once again $h^* \succ h$. Thus, for a sample in ${\cal A}$, $h^* \succ h$ for every $h \in H$, implying that $h^* \in \mP_2(H)$.

To prove the second part, consider any $h \in \mP_2(H)$, and in particular, $h \succ h^*$. Observe that $\B_{h^*,h} = - \B_{h,h^*}$, and since $h \succ h^*$ one has that for more than $n/2$ of the coordinate blocks,
\begin{equation*}
\begin{cases}
\B_{h,h^*}(j) \leq \left((\beta/\alpha)^2+\gamma\right)r^2 & \mbox{if} \ \ \ \mP_1(h,f) < \beta r,
\\
\\
\B_{h,h^*}(j) \leq \left(2\nu/\alpha^2\right) \mP_1^2(h,f) & \mbox{if } \ \ \ \mP_1(h,f) \geq \beta r.
\end{cases}
\end{equation*}
Examining the two possibilities, if $\mP_1(h,f) < \beta r$ then by Condition $(1)$, $\|h-h^*\|_{L_2} \leq (\beta/\alpha) r$. Also, combining \eqref{eq:B-in-proof} and Condition $(3)$, there is a coordinate block $j$ on which both
\begin{align*}
\B_{h,h^*}(j) = & \E{\cal L}_h +  \left(\Q_{h,h^*}(j)-\|h-h^*\|_{L_2}^2\right) + \left(\M_{h,h^*}(j) - \E\M_{h,h^*}(j)\right)
\\
\leq & \left((\beta/\alpha)^2+\gamma\right)r^2
\end{align*}
and
$$
|\M_{h,f}(j)-\E \M_{h,f} | \leq \gamma r^2.
$$
For that block,
\begin{align*}
\E{\cal L}_h \leq & - \left(\Q_{h,h^*}(j)-\|h-h^*\|_{L_2}^2\right) - \left(\M_{h,h^*}(j) - \E\M_{h,h^*}\right) + \left((\beta/\alpha)^2+\gamma \right)r^2
\\
\leq & \|h-h^*\|_{L_2}^2 + |\M_{h,h^*}(j) - \E \M_{h,h^*}| + ((\beta/\alpha)^2+\gamma)r^2
\\
\leq & 2\left((\beta/\alpha)^2+\gamma \right)r^2
\end{align*}
and $h$ satisfies \eqref{eq:observation}.

If, on the other hand, $\mP_1(h,h^*) \geq \beta r$ then by Condition $(1)$, 
$$
\alpha \|h-h^*\|_{L_2} \leq \mP_1(h,h^*) \leq \beta \|h-h^*\|_{L_2};
$$
thus, $\|h-h^*\|_{L_2} \geq r$ and for more than $n/2$ of the coordinate blocks, 
\begin{equation} \label{eq:in-proof-B-1}
\B_{h,h^*}(j)
\leq 2\frac{\nu}{\alpha^2} \mP_1^2(h,h^*) \leq 2\nu \frac{\beta^2}{\alpha^2} \|h-h^*\|_{L_2}^2,
\end{equation}
Combining that with Condition $(2)$, there is a coordinate block $j$ on which both \eqref{eq:in-proof-B-1} and
$$
\Q_{h,h^*}(j) \geq (1-\nu)\|h-h^*\|_{L_2}^2,  \ \ \M_{h,h^*} - \E \M_{h,h^*} \geq -\nu \|h-h^*\|_{L_2}^2
$$
hold. Moreover, by \eqref{eq:B-in-proof},
$$
\B_{h,h^*}(j) = \E{\cal L}_h +  \left(\Q_{h,h^*}(j)-\|h-h^*\|_{L_2}^2\right) + \left(\M_{h,h^*}(j) - \E\M_{h,h^*}(j)\right)
$$
implying that
$$
\E{\cal L}_h \leq 2\nu\left(1+\frac{\beta^2}{\alpha^2}\right) \|h-h^*\|_{L_2}^2,
$$
as required in \eqref{eq:observation}.
\endproof

Finally, we are ready for the `main event': showing that the random components of Theorem \ref{thm:observation} are indeed true; and specifying the tuning parameters $\ell,\theta_1,...,\theta_4$ that are needed for the definition of the procedure $\mP$ in terms of the uniform integrability function $\kappa(\xi)$.

\subsection{Proof of the random components} \label{sec:proof-rand}
The random components are all based on the same requirement: that a certain property (denoted by {\bf P}) holds for any function in the class (or in an appropriate subset of a class) on a majority of the coordinate blocks. The way one obtains such a uniform estimate is at the heart of the small-ball method:

\begin{description}
\item{$(a)$} Show that with very high probability, {\bf P} holds for a single function on a significant majority of the coordinate blocks---by first proving {\bf P} for a single function and a single block and then passing from a single block to a significant majority of blocks using a binomial estimate.

\item{$(b)$} The very high probability in $(a)$ and the union bound allows one to extend the claim from a single function to a relatively large (finite) collection of functions, all satisfying {\bf P} on a significant majority of the coordinate blocks. The finite collection one selects is an appropriate net in the given class.

\item{$(c)$} Obtain a uniform bound on the `oscillations': a high probability event on which, for \emph{any} pair of functions $f,h$ in the class that are close enough, $h-f$ cannot ruin property {\bf P} on too many coordinates blocks.

\item{$(d)$} Finally, consider the intersection of the events from $(b)$ and $(c)$. For any $h$, let $f$ be a function in the net that is sufficiently close to $h$. By $(b)$, $f$ satisfies property {\bf P} on a significant majority of the blocks, while $(c)$ implies that $h-f$ does not ruin property {\bf P} on too many coordinate blocks. Thus, $h$ satisfies property {\bf P} on a (smaller) majority of the blocks.
\end{description}

The claims we establish in this way are an almost isometric lower bound on $\Q_{h,h^*}(j)$; an isomorphic two-sided bound on $\Q_{h,h^*}(j)$; and an upper bound on $|\M_{h,h^*}(j)-\E \M_{h,h^*}|$, all of which hold on a significant majority of the coordinate blocks and for every $h$ in an appropriate large subset of the class.

\vskip0.3cm

\subsubsection*{Almost isometric lower estimates on $\Q$}
The first probabilistic component we present is an almost isometric lower bound on quadratic forms: that for any $0<\xi<1/4$, with high probability, if $h \in H$ and $\|h-h^*\|_{L_2}$ is large enough then  $\Q_{h,h^*}(j) \geq (1-4\xi)\|h-h^*\|_{L_2}^2$ on a significant majority of the coordinate blocks. Although the class we are interested in is $H-h^*$, to ease notation we set $h^*=0$; thus the class $H_{h^*,r}$ becomes $H_r={\rm star}(H,0) \cap rD$.

\vskip0.3cm

The starting point is the following property, denoted by {\bf P1}:
\begin{Definition}
For constants $0<\xi<1/4$ and $1 \leq \ell \leq m$, the function $h$ satisfies property {\bf P1} on $I=\{1,...,m\}$ if for any $J \subset \{1,...,m\}$ of cardinality $|J| \leq \ell$,
$$
\frac{1}{m}\sum_{i \in J^c} h^2(X_i) \geq (1-3\xi)\|h\|_{L_2}^2.
$$
\end{Definition}
The idea behind {\bf P1} is geometric: one should show that the random vector $(|h(X_i)|)_{i=1}^N$ is `well-spread', and thanks to that, \emph{an almost isometric lower estimate is stable}: it is satisfied even if a relatively small subset of the coordinates is excluded from the sum. Naturally, the highest impact on the sum occurs when removing the largest $\ell$ coordinates of $(|h(X_i)|)_{i=1}^N$, and {\bf P1} implies that the impact those coordinates have on the sum is relatively negligible.

\vskip0.3cm

We begin by showing that a single function satisfies {\bf P1} with high probability.
\begin{Lemma} \label{lemma:lower-single}
For a function $h$ and $0<\xi<1/3$, let $\kappa=\kappa(\xi)$ satisfy that $\E h^2 \IND_{\{|h| \geq \kappa\|h\|_{L_2}\}} \leq \xi \|h\|_{L_2}^2$. If $\ell=m \xi/\kappa^2$, then $h$ satisfies property {\bf P1} with probability at least $1-2\exp(-cm\xi^2/\kappa^2)$, where $c$ is an absolute constant. In other words, with that probability, for every $J \subset \{1,...,m\}$ of cardinality $|J| \leq \ell$,
$$
\frac{1}{m}\sum_{i \in J^c} h^2(X_i) \geq (1-3\xi)\|h\|_{L_2}^2.
$$
\end{Lemma}

\begin{Remark}
Note that the condition on $h$ follows from the uniform integrability condition appearing in Assumption \ref{ass:main-on-F}.
\end{Remark}

\proof Let $f=h^2 \IND_{\{|h| \leq \kappa\|h\|_{L_2}\}}$ and observe that $\E f \geq (1-\xi)\|h\|_{L_2}^2$, $\|f\|_{L_\infty} \leq (\kappa\|h\|_{L_2})^2$, and $\E f^2 \leq \|f\|_{L_\infty} \E h^2 \leq \kappa^2 \|h\|_{L_2}^4$. Applying Bernstein's inequality,
$$
Pr \left( \left|\frac{1}{m}\sum_{i=1}^m f(X_i)-\E f \right| \geq \xi \|h\|_{L_2}^2 \right) \leq 2 \exp\left(-c\frac{m \xi^2}{\kappa^2}\right).
$$
Hence, for every $J \subset \{1,...,m\}$
\begin{align*}
\frac{1}{m}\sum_{i \in J^c} h^2(X_i) \geq & \frac{1}{m} \sum_{i \in J^c} f(X_i) \geq \E f - \left|\frac{1}{m}\sum_{i=1}^m f(X_i) - \E f \right| - \frac{|J|}{m} \cdot \|f\|_{L_\infty}
\\
\geq & (1-2\xi) \|h\|_{L_2}^2 - \frac{|J|}{m} \kappa^2 \|h\|_{L_2}^2 \geq (1-3\xi)\|h\|_{L_2}^2,
\end{align*}
provided that $|J| \leq m\xi/\kappa^2$.
\endproof
The next step is to pass from {\bf P1} on a single block to {\bf P1} on a majority of the blocks.
\begin{Corollary} \label{cor:proof-lower-1}
There exist absolute constants $c_0$ and $c_1$ such that the following holds. Fix any $0<\xi<1/3$ and set $\kappa=\kappa(\xi)$. Let $(I_j)_{j=1}^n$ be the partition of $\{1,...,N\}$ to $n$ coordinate blocks of cardinality $m$. If $0<\eta<1/2$ and
\begin{equation} \label{eq:in-cor-m}
m \geq c_0 \left(\frac{\kappa^2}{\xi^2}\right) \log \left(\frac{2}{\eta}\right)
\end{equation}
then with probability at least $1-2\exp(-c_1N \eta \xi^2/\kappa^2)$, $h$ satisfies {\bf P1} on at least $(1-\eta)n$ coordinate blocks.
\end{Corollary}

To put Corollary \ref{cor:proof-lower-1} in some perspective, recall that by \eqref{eq:cond-on-N} and \eqref{eq:condition-on-m},
$$
N \geq \theta_0 \log \left(\frac{64}{\delta}\right) \ \ \ \ {\rm and } \ \ \ \ n = \theta_1 \log \left(\frac{64}{\delta}\right);
$$
therefore, the number of coordinates in each block is at least
$$
m \geq \frac{\theta_0}{\theta_1}.
$$
Invoking \eqref{eq:in-cor-m}, if
\begin{equation} \label{eq:cond-on-theta-1}
\theta_0 \geq \frac{\kappa^2}{\xi^2} \max\left\{c_0 \theta_1 \log \left(\frac{2}{\eta}\right),\frac{1}{c_1\eta}\right\},
\end{equation}
then the assertion of Corollary \ref{cor:proof-lower-1} holds with probability at least
$$
1-2\exp\left(-c_1N \eta \frac{\xi^2}{\kappa^2}\right) \geq 1-\frac{\delta}{32}.
$$

\vskip0.4cm

\proof Let $\sharp(h)$ be the number of coordinate blocks $I_j$ on which $h$ satisfies property {\bf P1}. Lemma \ref{lemma:lower-single} implies that for a given coordinate block $I_j$, $h$ satisfies {\bf P1} with probability at least $1-2\exp(-cm\xi^2/\kappa^2)$. Let $\zeta=2\exp(-cm\xi^2/\kappa^2)$, set $(\zeta_i)_{i=1}^n$ to be independent $\{0,1\}$-valued random variables with mean $\zeta$, and note that by \eqref{eq:in-cor-m}, $2\zeta < \eta \leq 1/2$. A straightforward application of Bennett's inequality reveals that
$$
\sum_{i=1}^n \zeta_i \leq \eta n
$$
with probability at least
$$
1-2\exp(-c^\prime \eta n \log(\eta/\zeta)) \geq 1-2\exp\left(-c^\prime \eta n \left( cm\xi^2/\kappa^2 - \log(1/\eta) \right) \right);
$$
Hence, with that probability, $h$ satisfies property {\bf P1} on at least $(1-\eta)n$ of the coordinate blocks $I_j$. Also, since
\begin{equation} \label{eq:proof-cond-1}
cm\xi^2/\kappa^2 \geq 2\log(1/\eta),
\end{equation}
and $nm=N$, it is evident that
\begin{equation} \label{eq:Prop1-and probability}
\sharp(h) \geq (1-\eta)n \ \ {\rm with \ probability \ \ } 1-2\exp\left(-c^{\prime \prime} N \eta \xi^2/\kappa^2 \right)
\end{equation}
for a suitable absolute constant $c^{\prime \prime}$.
\endproof


The final step is passing from a high probability estimate for a single function to a uniform estimate for a class of functions; that is, given a class $H$, to show that with high probability, if $\|h\|_{L_2}$ is larger than some threshold, then
$$
\frac{1}{m}\sum_{i \in I_j} h^2(X_i) \geq (1-4\xi)\|h\|_{L_2}^2
$$
for at least $(1-2\eta)n$ of the coordinate blocks $I_j$. An equivalent formulation of that fact is that for a given threshold one requires the sample size to be large enough --- and with our choice of $N$ that will turn out to be the case.

As expected, the threshold in question is given in terms of fixed points that capture the `local structure' of the class around $h^*$; and since we set $h^*=0$, this corresponds to the local structure of the star-shaped hull ${\rm star}(H,0)$ around $0$.

\vskip0.4cm

For any $0<\xi<1/4$, $\kappa=\kappa(\xi)$ and $0<\eta<1/4$, let $\gamma_1,...,\gamma_3$ be constants that depend only on $\xi$, $\kappa$ and $\eta$ and that will be specified in what follows. Consider the sets $H_r={\rm star}(H,0) \cap rD$ and let $r$ satisfy that
\begin{equation} \label{eq:in-proof-uniform-lower-1}
\log {\cal M}(H_r, \gamma_1r D) \leq \gamma_2N \ \ \  {\rm and} \ \ \  \E\sup_{h \in H_r} |\sum_{i=1}^N \eps_i h(X_i)| \leq \gamma_3r N.
\end{equation}

\begin{Theorem} \label{thm:uniform-lower-isometric}
Set $\gamma_1,\gamma_2,\gamma_3 \lesssim \frac{\eta \xi^2}{\kappa^2(\xi)}$, $c \sim \eta^2$, recall that $m \geq c_0(\kappa^2(\xi)/\xi^2) \log (2/\eta)$ as in Corollary \ref{cor:proof-lower-1} and that $n=N/m$. Then, with probability at least $1-2\exp(-c n)$, for any $h \in H$ such that $\|h\|_{L_2} \geq r$ there are at least $(1-2\eta)n$ coordinate blocks $I_j$ on which
\begin{equation*}
\frac{1}{m}\sum_{i \in I_j} h^2(X_i) \geq (1-4\xi)\|h\|_{L_2}^2.
\end{equation*}
\end{Theorem}

\begin{framed}
Taking into account \eqref{eq:cond-on-theta-1}, the assertion of Theorem \ref{thm:uniform-lower-isometric} holds with probability at least $1-\delta/16$ if
$$
\theta_1 \gtrsim \frac{1}{\eta^2}, \ \ \ \ {\rm and} \ \ \ \ \theta_0 \gtrsim \frac{\kappa^2}{\xi^2} \max \left\{\theta_1 \log \left(\frac{2}{\eta}\right), \frac{1}{\eta} \right\}.
$$
\end{framed}

\proof
Fix $0<\xi<1/4$ and set $\kappa=\kappa(\xi)$. Let $H^\prime$ be a maximal $\gamma_1 r$-separated subset in ${\rm star}(H,0) \cap rS \subset H_r$ and assume that $\log {\cal M}(H_r, \gamma_1r) \leq \gamma_2N$ for $\gamma_2 \sim \eta \xi^2/\kappa^2$. Let ${\cal A}_1$ be the event
\begin{equation} \label{eq:high-probab-S}
\sharp(h^\prime) \geq (1-\eta)n \ \ \ {\rm for \ every \ } h^\prime \in H^\prime,
\end{equation}
and observe that by Corollary \ref{cor:proof-lower-1} combined with the union bound,
$$
Pr({\cal A}_1) \geq 1-2\exp\left(-cN \eta \frac{\xi^2}{\kappa^2}\right).
$$

Next, consider the `oscillations' $h-\pi h$, where $h \in {\rm star}(H,0) \cap rS$ and $\pi h \in H^\prime$ satisfies that $\|h-\pi h\|_{L_2} \leq \gamma_1 r$. The crucial point in the argument is that the differences $h-\pi h$ do not ruin {\bf P1} on too many coordinate blocks. To express that formally, for any $h_1,h_2 \in {\rm star}(H,0) \cap rS$ set
$$
R_j(h_1,h_2)=|\{i \in I_j: |h_1-h_2|(X_i) \geq \xi r\}|
$$
and let ${\cal A}_2$ be the event defined by
\begin{equation} \label{eq:osc-quad}
\sup_{h \in {\rm star}(H,0) \cap rS} \frac{1}{n}\sum_{j=1}^n \IND_{\{R_j(h,\pi h) \geq \ell\}} \leq \eta.
\end{equation}
On ${\cal A}_2$, for every $h \in {\rm star}(H,0) \cap rS$ there are at most $\eta n$ coordinate blocks $I_j$ where $|h(X_i)-\pi h(X_i)| \geq \xi r$ for more than $\ell$ coordinates in the block $I_j$. Therefore, if $(X_i)_{i=1}^N \in {\cal A}_1 \cap {\cal A}_2$ then:
\begin{description}
\item{$(1)$} Every $\pi h$ satisfies property {\bf P1} on at least $(1-\eta)n$ coordinate blocks $I_j$. For those $I_j$'s and for any $J_j \subset I_j$ of cardinality at most $\ell=m\xi/\kappa^2$,
\begin{equation} \label{eq:in-proof-LQ1}
\frac{1}{m}\sum_{i \in I_j \backslash J_j} (\pi h)^2(X_i) \geq (1-3\xi)\|\pi h\|_{L_2}^2 = (1-3\xi)r^2.
\end{equation}
\item{$(2)$} By \eqref{eq:osc-quad}, there are at most $\eta n$ coordinate blocks on which $|h-\pi h|(X_i) \geq \xi r$ on more than $\ell$ coordinates.
\end{description}
Hence, there are at least $(1-2\eta)n$ coordinate blocks on which both properties are satisfied. For each one of those blocks $I_j$, one may set $J_j =\{i \in I_j : |h-\pi h|(X_i) \geq \xi r\}$ and \eqref{eq:in-proof-LQ1} holds. Also, as $|h-\pi h|(X_i) \leq \xi r$ for $i \in I_j \backslash J_j$, it is evident that
\begin{align*}
\Bigl(\frac{1}{m}\sum_{i \in I_j} h^2(X_i)\Bigr)^{1/2} \geq & \Bigl(\frac{1}{m}\sum_{i \in I_j \backslash J_j} h^2(X_i)\Bigr)^{1/2}
\\
\geq &
\Bigl(\frac{1}{m}\sum_{i \in I_j \backslash J_j} (\pi h)^2(X_i)\Bigr)^{1/2} - \Bigl(\frac{1}{m}\sum_{i \in I_j \backslash J_j} (h-\pi h)^2(X_i)\Bigr)^{1/2}
\\
\geq & (1-3\xi)^{1/2}r - \xi r.
\end{align*}
Hence, on at least $(1-2\eta)n$ coordinate blocks $I_j$,
\begin{equation} \label{eq:quadratic-level-r}
\frac{1}{m}\sum_{i \in I_j} h^2(X_i) \geq (1-4\xi)\|h\|_{L_2}^2.
\end{equation}
The same estimate is true for any $h \in H$ that satisfies $\|h\|_{L_2} \geq r$ because ${\rm star}(H,0)$ is star-shaped around $0$ and \eqref{eq:quadratic-level-r} is positive homogeneous.
\vskip0.3cm
All that is left to complete the proof of Theorem \ref{thm:uniform-lower-isometric} is to estimate $Pr({\cal A}_2)$ and specify the restriction on the constants $\gamma_1$ and $\gamma_3$. To that end, observe that
\begin{equation*}
\sum_{j=1}^n \IND_{\{R_j(h,\pi h) \geq \ell\}}  \leq \frac{1}{\ell} \sum_{j=1}^n R_j(h,\pi h)= \frac{1}{\ell} \sum_{i=1}^n \sum_{i \in I_j} \IND_{\{|h-\pi h|(X_i) \geq \xi r\}} \leq \frac{1}{\ell \xi r} \sum_{i=1}^N |h-\pi h|(X_i).
\end{equation*}
By the Gin\'{e}-Zinn symmetrization Theorem \cite{MR757767},
\begin{align*}
& \E \sup_{h \in {\rm star}(H,0) \cap rS} \sum_{j=1}^n \IND_{\{R_j(h,\pi h) \geq \ell\}}
\\
\leq & \frac{1}{\ell \xi r} \left( \E \sup_{h \in {\rm star}(H,0) \cap rS}\sum_{i=1}^N \left(|h-\pi h|(X_i) - \E|h-\pi h|(X_i) \right) + N\sup_{h \in {\rm star}(H,0) \cap rS} \E|h-\pi h|(X_i) \right)
\\
\leq &  \frac{2}{\ell \xi r} \E \sup_{h \in {\rm star}(H,0) \cap rS} \left|\sum_{i=1}^N \eps_i (h-\pi h) (X_i)\right| + \frac{N}{\ell \xi r}\sup_{h \in {\rm star}(H,0) \cap rS} \|h-\pi h\|_{L_2}
\\
\leq & \frac{4m}{\ell \xi r} \cdot \frac{1}{m} \E \sup_{h \in {\rm star}(H,0) \cap rS} \left|\sum_{i=1}^N \eps_i h(X_i)\right| + \frac{N \gamma_1}{\ell \xi }.
\end{align*}
Therefore, if
$$
Z=\sup_{h \in {\rm star}(H,0) \cap rS} \frac{1}{n}\sum_{j=1}^n \IND_{\{R_j(h,\pi h) \geq \ell\}},
$$
then
$$
\E Z \leq \frac{4m}{\ell \xi r} \frac{1}{N} \E \sup_{h \in {\rm star}(H,0) \cap rS} \left|\sum_{i=1}^N \eps_i h(X_i)\right| + \frac{m \gamma_1}{\ell \xi}=(1)+(2).
$$
To ensure that $\E Z \leq \eta/2$ it suffices that $(1),(2) \leq \eta/4$, i.e.,
\begin{equation} \label{eq:inproof-cond-on-r-1}
\frac{1}{N} \E \sup_{h \in H_r} \left|\sum_{i=1}^N \eps_i h(X_i)\right| \leq r \cdot  \frac{\eta \ell \xi}{16 m} = r  \cdot \frac{\eta \xi^2}{16 \kappa^2}
\end{equation}
and
\begin{equation} \label{eq:inproof-cond-on-r-1a}
\gamma_1 \leq  \frac{\eta \ell \xi}{4m} =  \frac{\eta \xi^2}{4\kappa^2};
\end{equation}
both can be verified with the choices of $\gamma_1,\gamma_3 \lesssim \eta \xi^2/\kappa^2$.

Finally, by the bounded differences inequality (see, e.g., \cite{BoLuMa13}),
$$
Pr(Z \geq \E Z + u/\sqrt{n}) \leq \exp(-c u^2),
$$
implying that $Pr({\cal A}_2) \geq 1-\exp(-c\eta^2 n)$, as claimed.
\endproof

In what follows we apply Theorem \ref{thm:uniform-lower-isometric} twice: first to obtain a two-sided isomorphic bound and then for an almost isometric lower bound.

\subsubsection*{A two-sided isomorphic bound}
Again, as in the previous section we ease notation by setting $h^*=0$. 

The two-sided isomorphic estimate can be derived from a lower one because the upper estimate is, to a certain extent, universally true. To formulate the claim, denote by $(h(X_i))_\ell^*$ the $\ell$-largest value in a monotone rearrangement of the coordinates of the vector $(|h(X_i)|)_{i=1}^N$. In what follows we set
$$
\xi_1 = \frac{1}{10}; \ \ \ \kappa_1 = \max\{\kappa(\xi_1),1\}; \ \ {\rm and} \ \ \eta=\frac{\xi_1}{\kappa_1^2} = \frac{1}{10 \kappa_1^2}.
$$

\begin{Theorem} \label{thm:iso-two-sided}
There exist absolute constants $c_1,c_2,c_3,c_4$ and $c_5$ for which the following holds. Assume that for every $h \in {\rm star}(H,0)$, $\E h^2\IND_{\{|h| \geq \kappa(\xi_1) \|h\|_{L_2}\}} \leq \xi_1 \|h\|_{L_2}^2$. Let $H_r=  {\rm star}(H,0) \cap rD$ for a radius $r$ such that
\begin{equation} \label{eq:in-thm-iso-two-sided}
\E \sup_{h \in H_r} \left|\frac{1}{\sqrt{N}}\sum_{i=1}^N \eps_i h(X_i)\right| \leq \frac{c_1 r}{\kappa_1^2} \sqrt{N} \ \ {\rm and} \ \ \log{\cal M}\left(H_r, \frac{c_2}{\kappa_1^2} D\right) \leq \frac{c_3}{\kappa_1^2} N.
\end{equation}
Then, on an event with probability at least $1-2\exp(-c_4N/\kappa_1^4)$,
\begin{description}
\item{$\bullet$} if $\|h\|_{L_2} \geq r$ then $(1/2\sqrt{10})\|h\|_{L_2} \leq (h(X_i))_{N/5\kappa_1^2}^* \leq 3\kappa_1 \|h\|_{L_2}$, and
\item{$\bullet$} if $\|h\|_{L_2} < r$ then $(h(X_i))_{N/5\kappa_1^2}^* \leq 3\kappa_1 r$.
\end{description}
\end{Theorem}

\begin{framed}
Recall that $N \geq \theta_0 \log(64/\delta)$. Thus, the claim of Theorem \ref{thm:iso-two-sided} holds with probability at least $1-\delta/32$  provided that
\begin{equation} \label{eq:cond-on-theta-2}
\theta_0 \geq \frac{\kappa_1^4}{c_4} \sim \max\{\kappa^4(1/10),1\}.
\end{equation}
\end{framed}
\vskip0.4cm
As we did previously, let us specify the feature of a single function we require.
\begin{Definition} \label{def:P2}
Given a sample $(X_i)_{i=1}^N$ and a function $h$, set $v=(|h(X_i)|)_{i=1}^N$. For a fixed $0<\xi <1$, $h$ satisfies {\bf P2} on the sample if, setting $\ell=\xi N/\kappa^2(\xi)$ and $s=N(1-4\xi)/(2\kappa^2(\xi))$, one has
$$
\sqrt{\xi} \|h\|_{L_2} \leq v_s^* \leq v_\ell^* \leq 2\kappa(\xi) \|h\|_{L_2}.
$$
\end{Definition}
In other words, {\bf P2} means that $\sim N$ of the values $|h(X_i)|$ are proportional to $\|h\|_{L_2}$, with constants that depend only on the uniform integrability estimate that $h$ satisfies.

\begin{Lemma} \label{Lemma-in-proof-two-sided-single}
If $\E h^2\IND_{\{|h| \geq \kappa \|h\|_{L_2}\}} \leq \xi \|h\|_{L_2}^2$ then $h$ satisfies {\bf P2} with probability at least $1-4\exp(-cN\xi^2/\kappa^2)$.
\end{Lemma}

\proof
By Lemma \ref{lemma:lower-single} applied for $m=N$ and $\kappa=\kappa(\xi)$, we have that with probability at least
$1-2\exp(-cN\xi^2/\kappa^2)$, for any $J \subset \{1,...,N\}$ of cardinality $|J| \leq N \xi/\kappa^2$,
\begin{equation} \label{eq:in-proof-isomorphic-single-lower}
\frac{1}{N}\sum_{i \in J^c} h^2(X_i) \geq (1-3\xi)\|h\|_{L_2}^2;
\end{equation}
and, by Chebychev's inequality, $Pr(|h| \geq 2\kappa \|h\|_{L_2}) \leq \xi/4\kappa^2$. Therefore, a standard binomial estimate shows that with probability at least
$1-2\exp(-cN\xi/\kappa^2)$,
$$
\left|\left\{ i: |h(X_i)| \geq 2\kappa \|h\|_{L_2} \right\} \right| \leq \frac{\xi N}{\kappa^2}.
$$
On the intersection of the two events, and for $J \subset \{1,...,N\}$ consisting of the largest $\xi N/\kappa^2$ coordinates of $(|h(X_i)|)_{i=1}^N$,
$$
\frac{1}{N} \sum_{i \in J^c} h^2(X_i) \geq (1-3\xi)\|h\|_{L_2}^2 \ \ {\rm and} \ \ \max_{i \in J^c} |h(X_i)| \leq 2 \kappa \|h\|_{L_2}.
$$
The rest of the proof is a standard Paley-Zygmund type argument: set $0 < \lambda < 1$ and put $|\{i \in J^c : |h(X_i)| \geq \lambda \|h\|_{L_2}\}|=\theta N$. Hence,
$$
(1-3\xi)\|h\|_{L_2}^2 N \leq \sum_{i \in J^c} h^2(X_i) \leq \lambda^2 \|h\|_{L_2}^2 N + \theta N \cdot 2 \kappa^2 \|h\|_{L_2}^2
$$
and for $\lambda^2 \leq \xi$,
\begin{equation} \label{eq:in-proof-isomorphic-single-proportion}
\theta \geq \frac{1-4\xi}{2\kappa^2}.
\end{equation}
Therefore, the $(1-4\xi)N/(2\kappa^2)$ largest coordinates of $(|h(X_i)|)_{i \in J^c}$ are in the required range, as claimed.
\endproof

\noindent{\bf Proof of Theorem \ref{thm:iso-two-sided}.}
 Recall that we set $\xi_1=1/10$, $\kappa_1=\max\{\kappa(1/10),1\}$ and $\eta=\xi_1/\kappa_1$. Let $\alpha=\sqrt{\xi_1/2}$ and set $H^\prime$ be an $(\eta \alpha/8) r$ maximal separated set of ${\rm star}(H,0) \cap rS$ for a radius $r$ that satisfies \eqref{eq:in-thm-iso-two-sided}.

 to be specified in what follows. For every $h \in {\rm star}(H,0) \cap rS$, let $\pi h \in H^\prime$ be the closest to $h$ in the net, and in particular, $\|h-\pi h\|_{L_2} \leq (\eta \alpha/8)r$.

Note that if $c$ is a well-chosen absolute constant and
\begin{equation} \label{eq:entopy-cond-1}
\log {\cal M}\left(H_r, \frac{\alpha \eta}{8}rD \right) \leq cN \frac{\xi_1^2}{\kappa_1^2},
\end{equation}
then by the union bound, the assertion of Lemma \ref{Lemma-in-proof-two-sided-single} is true uniformly for any $\pi h \in H^\prime$ and with high probability. Specifically, if
$$
{\cal A}_1 = \left\{ {\rm every \ } \pi h \in H^\prime \ {\rm satisfies \ {\bf P2} } \right\},
$$
then $Pr({\cal A}_1) \geq 1-2\exp(-cN\xi_1^2/\kappa_1^2)$.

The heart of the proof is to show that with high probability,
\begin{equation} \label{eq:in-proof-isomorphic-oscillation-1}
Z= \sup_{h \in {\rm star}(H,0) \cap rS} \left|\left\{i : |(h - \pi h)(X_i)| \geq \frac{\alpha r}{2}\right\}\right| \leq \eta N.
\end{equation}
Indeed, for every $h \in {\rm star}(H,0) \cap rS$,
$$
\left|\left\{i : |(h - \pi h)(X_i)| \geq \frac{\alpha r}{2}\right\}\right| \leq \frac{2}{\alpha r} \sum_{i=1}^N |h-\pi h|(X_i)
$$
and therefore,
\begin{align*}
\E Z = & \E \sup_{h \in {\rm star}(H,0) \cap rS} \sum_{i=1}^N \IND_{\{ |h-\pi h| \geq \alpha r/2\}}
\\
\leq & \frac{2}{\alpha r} \E \sup_{h \in {\rm star}(H,0) \cap rD} \left|\sum_{i=1}^N \left(|h-\pi h|(X_i) - \E |h-\pi h| \right) \right| + \frac{2N}{\alpha r} \sup_{h \in {\rm star}(H,0) \cap rS} \|h-\pi h\|_{L_2}
\\
= & (1)+(2).
\end{align*}
To ensure that $\E Z \leq \eta N/2$ it suffices to verify that $(1),(2) \leq \eta N /4$. The latter is true because $\|h-\pi h\|_{L_2} \leq (\eta \alpha /8)r$; the former can be verified using symmetrization, followed by the triangle inequality and de-symmetrization, implying that
$$
(1) \leq \frac{8}{\alpha r} \E \sup_{h \in {\rm star}(H,0) \cap rD} \left|\frac{1}{\sqrt{N}} \sum_{i=1}^N \eps_i h(X_i) \right|.
$$
Hence, if
\begin{equation} \label{eq:osc-cond-1}
\E \sup_{h \in H_r} \left|\frac{1}{\sqrt{N}} \sum_{i=1}^N \eps_i h(X_i) \right| \leq \frac{ \eta \alpha}{32} \sqrt{N}r
\end{equation}
then $(1) \leq \eta N/4$.

Both \eqref{eq:entopy-cond-1} and \eqref{eq:osc-cond-1} hold by our choice of constants; thus $\E Z \leq \eta N/2$, and by the bounded differences inequality,
$$
Pr \left(Z \geq \eta N\right) \leq 2\exp(-c_1 \eta^2 N)
$$
for an absolute constant $c_1$.
\vskip0.3cm
Let ${\cal A}_2$ be the event given by \eqref{eq:in-proof-isomorphic-oscillation-1}. Observe that
$Pr({\cal A}_1 \cap {\cal A}_2) \geq 1-2\exp(-cN\xi_1^2/\kappa_1^4)$ and for $(X_i)_{i=1}^N \in {\cal A}_1 \cap {\cal A}_2$ and $h \in {\rm star}(H,0) \cap rS$:
\begin{description}
\item{$(a)$} if $v=(|\pi h|(X_i))_{i=1}^N$ then  $\sqrt{\xi_1} r \leq v_s^* \leq v_\ell^* \leq 2\kappa_1 r$ where $\ell=\xi_1 N/\kappa_1^2$ and $s=N(1-4\xi_1)/(2\kappa_1^2)$;
\item{$(b)$} $|(h - \pi h)(X_i)| \leq \frac{\sqrt{\xi_1} r}{2}$ on at least $(1-\eta) N$ coordinates.
\end{description}
Fix $h \in {\rm star}(H,0) \cap rS$ and consider an index $i$ on which both $(a)$ and $(b)$ occur. For such an index,
\begin{equation} \label{eq:in-proof-isomorphic-oscillation-2}
|h(X_i)| \leq |\pi h|(X_i)+|h-\pi h|(X_i) \leq (2\kappa_1  + \sqrt{\xi_1}/2) r
\end{equation}
and
\begin{equation} \label{eq:in-proof-isomorphic-oscillation-3}
|h(X_i)| \geq |\pi h|(X_i)-|h-\pi h|(X_i) \geq \frac{\sqrt{\xi_1} r}{2}.
\end{equation}
Note that \eqref{eq:in-proof-isomorphic-oscillation-2} is true for every coordinate $i$ that is not one of the $\xi_1 N/\kappa_1^2$ largest coordinates of $(|\pi h(X_i)|)_{i=1}^N$ and also not one of the $\eta N$ largest coordinates of $(|h-\pi h|(X_i))_{i=1}^N$. Hence, if $u_i=h(X_i)$ and $\ell_1=(\eta+\xi_1/\kappa_1^2)N$ then $u^*_{\ell_1} \leq (2\kappa_1  + \sqrt{\xi_1}/2) r$. Moreover, \eqref{eq:in-proof-isomorphic-oscillation-3} is true on the subset of coordinates on which $(a)$ holds, excluding at most $\eta N$ coordinates of that set. There are at least $s_1=s-\ell-\eta N$ such coordinates, and in particular, $u^*_{s_1} \geq \frac{\sqrt{\xi_1} r}{2}$. All that remains is to ensure that $\ell_1 < s_1$, i.e., that
$$
\frac{1-6\xi_1}{2\kappa_1^2} - \eta  \geq \eta+\frac{\xi_1}{\kappa_1^2},
$$
which can be verified for our choices $\xi_1=1/10$ and $\eta=\xi_1/\kappa_1^2$. Hence, $\ell_1=N/5\kappa_1^2$ and for $(X_i)_{i=1}^N \in {\cal A}_1 \cap {\cal A}_2$,
\begin{equation} \label{eq:in-proof-isomorphic}
\frac{r}{2 \sqrt{10}} \leq (h(X_i))_{N/5\kappa_1^2}^*  \leq 3\kappa_1 r.
\end{equation}
This resolves the case $\|h\|_{L_2}=r$; the assertion is automatically true for if $\|h\|_{L_2} > r$ because ${\rm star}(H,0)$ is star-shaped around $0$ and since \eqref{eq:in-proof-isomorphic} is positive homogeneous.
\vskip0.3cm

The proof of the second part, for functions in $H$ whose $L_2$ norm is smaller than $r$ follows the same path as the first one, by using {\bf P2} for a maximal separated subset of ${\rm star}(H,0) \cap rD$, (i.e., $\|\pi h\|_{L_2} \leq r$) followed by \eqref{eq:in-proof-isomorphic-oscillation-2}. We omit the standard details.
\endproof

\subsubsection*{A two-sided estimate on $\M$}

In what follows we set $\eta=0.01$, fix $0<\nu <1$ and let $\xi=\nu/4$.


Note that by the definition of $L_T$, for every $h \in H$
$$
\E (h-h^*)^2(X) \cdot (h^*(X)-Y)^2 \leq L_T^2 \E (h-h^*)^2 \cdot \E (h^*(X)-Y)^2,
$$
and thanks to homogeneity, $\E u^2(X) \cdot (h^*(X)-Y)^2 \leq L_T^2 \E u^2 \cdot \E (h^*(X)-Y)^2$ for any $u \in {\rm star}(H-h^*,0)$.

Finally, recall that
$$
\M_{h,h^*}(j)=\frac{2}{m} \sum_{i \in I_j} (h-h^*)(X_i) \cdot (h^*(X_i)-Y_i).
$$

\begin{Theorem} \label{thm:P3}
There exist absolute constants $c_1$ and $c_2$ for which the following holds.  Let
$$
\theta_1 \gtrsim 1 \ \ \ \ {\rm and} \ \ \ \ \theta_0 \geq \theta_1\left(\frac{16}{\eta \nu}\right)^2 ;
$$
set $\rho \geq r$ and assume that
$$
\E \sup_{u \in H_{h^*,\rho}} \left|\frac{1}{\sqrt{N}} \sum_{i=1}^N \eps_i (h^*(X_i)-Y_i) u(X_i) \right| \leq c_1 \nu \sqrt{N} \rho^2.
$$
Then with probability at least $1-2\exp(-c_2  n) \geq 1-\delta/32$, for every $h \in H$
\begin{description}
\item{$\bullet$} if $\|h-h^*\|_{L_2} \geq \rho$ then
$$
|\M_{h,h^*}(j)-\E \M_{h,h^*} | \leq \nu \|h-h^*\|_{L_2}^2 \ \ {\rm on \ at \ least \ } 0.99n \ {\rm coordinate \ blocks};
$$
\item{$\bullet$} if $\|h-h^*\|_{L_2} \leq \rho$ then
$$
|\M_{h,h^*}(j)-\E \M_{h,h^*} | \leq \nu \rho^2 \ \ {\rm on \ at \ least \ } 0.99n \ {\rm coordinate \ blocks}.
$$
\end{description}
\end{Theorem}

\proof
For every $h \in H$ let $u=h-h^*$ and set
$$
W_u(j) = \M_{h,h^*}(j)-\E \M_{h,h^*}=\frac{2}{m} \sum_{i \in I_j} u(X_i) \cdot (h^*(X_i)-Y_i)- \E u \cdot (h^*(X)-Y).
$$
Observe that $W_u$ is homogeneous in $u$ and let
$$
Z=\frac{1}{n}\sup_{u \in {\rm star}(H-h^*,0) \cap \rho S} |\{ j : |W_u(j)| \geq \nu \rho^2\} |.
$$
The aim is to show that with high probability, $Z \leq \eta=0.01$. Indeed,
\begin{align*}
Z \leq & \frac{1}{\nu \rho^2} \sup_{u \in {\rm star}(H-h^*) \cap \rho S}
\frac{1}{n}\sum_{j =1}^n |W_u(j)|
\\
\leq & \frac{1}{\nu \rho^2} \sup_{u \in H_{h^*,\rho}} \left| \frac{1}{n}\sum_{j=1}^n  |W_u(j)| - \E |W_u(j)| \right| + \frac{1}{\nu \rho^2} \sup_{u \in {\rm star}(H-h^*,0) \cap \rho S} \E |W_u|.
\end{align*}
Therefore, by symmetrization and contraction,
\begin{equation*}
\E Z \leq \frac{2}{\nu \rho^2} \E \sup_{u \in H_{h^*,\rho}} \left| \frac{1}{n} \sum_{j=1}^n  \eps_i W_u(j) \right| +  \frac{1}{\nu \rho^2} \sup_{u \in {\rm star}(H-h^*,0) \cap \rho S} \E |W_u| = (1)+(2).
\end{equation*}

To estimate $(2)$, note that
\begin{equation*}
\E |W_u| \leq  \frac{2}{m}
\E \left(\sum_{i \in I_j} u^2(X_i) \cdot (h^*(X_i)-Y_i)^2\right)^{1/2}
\leq \frac{2 L_T \sigma}{\sqrt{m}}  \|u\|_{L_2} \leq \frac{2 L_T \sigma \rho}{\sqrt{m}}.
\end{equation*}
Recall that $\|u\|_{L_2} = \rho >r $, and it follows from the definitions of $n$ and $N$ that
$$
\frac{1}{\sqrt{m}} = \sqrt{\frac{n}{N}} \leq \sqrt{\frac{\theta_1}{\theta_0}}\cdot \frac{r}{L_T \sigma}.
$$
Therefore,
\begin{equation} \label{eq:M-expectation}
\frac{2}{\nu \rho^2}  \E |W_u| \leq \frac{2L_T \sigma}{\nu \rho} \cdot \sqrt{\frac{n}{N}} \leq \frac{4}{\nu} \sqrt{\frac{\theta_1}{\theta_0}} \leq \frac{\eta}{4} = \frac{1}{400}
\end{equation}
provided that $\theta_0 \geq \theta_1 (16/\eta \nu)^2$, as was assumed.

Also,
\begin{equation*}
\E \sup_{u \in H_{h^*,\rho}} \left| \frac{1}{n}\sum_{j=1}^n  \eps_i W_u(j) \right|  \leq  2\E \sup_{u \in H_{h^*,\rho}} \left| \frac{1}{N} \sum_{i=1}^N \eps_i u(X_i) (h^*(X_i)-Y_i) \right|,
\end{equation*}
implying that
$$
(1) \leq \frac{4}{\nu \rho^2} \E \sup_{u \in H_{h^*,\rho}} \left| \frac{1}{N}\sum_{i=1}^N \eps_i u(X_i) (h^*(X_i)-Y_i) \right| \leq \frac{\eta}{4}
$$
when
\begin{equation} \label{eq:inproof-multi-cond1}
\E \sup_{u \in H_{h^*,\rho}} \left| \frac{1}{\sqrt{N}}\sum_{i=1}^N \eps_i u(X_i) (h^*(X_i)-Y_i) \right| \leq  \frac{\eta \nu}{16} \sqrt{N} \rho^2.
\end{equation}
Therefore, $\E Z \leq \eta/2$ and by the bounded differences inequality, $Pr(Z \geq \eta) \leq 2\exp(-c\eta^2 n)$, proving the first part of the theorem when $\|h-h^*\|_{L_2} = \rho$. When $\|h-h^*\|_{L_2} > \rho$, the first part holds by homogeneity in $h-h^*$.

The second part of the theorem, when $\|h-h^*\|_{L_2} \leq \rho$ is almost identical and we omit the straightforward details.
\endproof

\subsubsection*{Sudakov's inequality for conditional Bernoulli processes}
As was noted previously, our condition on the sample complexity $N$ was based on certain fixed points involving localized Rademacher averages. However, the proofs required an additional component: information on the packing numbers of localizations of the underlying class. Here, we show that the former implies the latter: that is, if $\gamma_3$ is small enough and \eqref{eq:Rad-Q-condition} holds, then
\begin{equation} \label{eq:entropy-condition}
\log {\cal M}(H_{h^*,r},\gamma_1 rD) \leq \gamma_2 N.
\end{equation}
Note that we may assume without loss of generality that $\gamma_2$ is sufficiently small, as the condition \eqref{eq:entropy-condition} becomes more restrictive the smaller $\gamma_2$ is.

\begin{Theorem} \label{thm:Bernoulli-implies-entropy}
Assume that there are constants $c_0$ and $c_1$ such that for every $w \in H-H$,
$$
Pr(|w| \geq c_0\|w\|_{L_2}) \geq c_1.
$$
If $\gamma_2 \leq c_1/8$ and $\log {\cal M}(H_{h^*,r},\gamma_1 rD) \geq \gamma_2 N$ then
$$
\E \sup_{u \in H_{h^*,r}} \left|\frac{1}{\sqrt{N}} \sum_{i=1}^N \eps_i u(X_i) \right| \geq \gamma_3 \sqrt{N} r
$$
provided that $\gamma_3 \leq c_2 \gamma_1 \sqrt{\gamma_2}$ for a constant $c_2$ that depends only on $c_0$ and $c_1$.
\end{Theorem}

In the context that interests us, the uniform integrability condition implies a small-ball condition, with $c_0$ and $c_1$ depending only on $\kappa(1/10)$. Hence, the outcome of Theorem \ref{thm:Bernoulli-implies-entropy} is that no matter what restrictions on $\gamma_1$ and $\gamma_2$ have been accumulated during the proof of Theorem \ref{thm:main}, by setting $\gamma_3 \leq c_2 \gamma_1 \sqrt{\gamma_2}$, \eqref{eq:entropy-condition} is automatically verified.

\begin{Remark} 
A more general version of Theorem \ref{thm:Bernoulli-implies-entropy} cab be found in \cite{Men:gen}.
\end{Remark}

\vskip0.3cm

The proof of Theorem \ref{thm:Bernoulli-implies-entropy} is based on Sudakov's inequality for Bernoulli processes \cite{LeTa91} in its scale-sensitive formulation (see, e.g.,  \cite{MR3274967}):
\begin{Theorem} \label{thm:Sudakov-for-Bernoulli}
There exists an absolute constant $c$ for which the following holds. Let $V \subset \R^N$ and for every $v \in V$ set $Z_v = \sum_{i=1}^N \eps_i v_i$. If $|V| \geq \exp(p)$ and $\{Z_v : v \in V\}$ is $\eps$-separated in $L_p$ then
$$
\E \sup_{v \in V} \sum_{i=1}^N \eps_i v_i \geq c \eps.
$$
\end{Theorem}

\noindent{\bf Proof of Theorem \ref{thm:Bernoulli-implies-entropy}.} Fix $u_1,u_2 \in H_{h^*,r}$. By the small-ball condition combined with a binomial estimate, we have that with probability at least $1-2\exp(-c_1N/2)$,
$$
|\{i: |u_1-u_2|(X_i) \geq c_0 \|u_1-u_2\|_{L_2} \}| \geq \frac{c_1 N}{4}.
$$
Applying the union bound, the same assertion holds with probability at least $1-2\exp(-c_1N/4)$ uniformly for every pair $u_k,u_\ell$ taken from a $\gamma_1 r$-separated subset of $H^\prime \subset H_{h^*,r}$, as long as $|H^\prime| \leq \exp(c_1N/8)$; such a subset exists if $\gamma_2 \leq c_1/8$, and in which case, its cardinality is $\exp(\gamma_2 N)$.

Consider any $u_k,u_\ell$ in the separated set, let $v=(u_k(X_i))_{i=1}^N$ and $w=(u_\ell(X_i))_{i=1}^N$ and put
$$
Z_{v}-Z_{w} = \sum_{i=1}^N \eps_i (v_i -w_i).
$$
By the characterization of the $L_p$ norm of the random variable $Z_a=\sum_{i=1}^N \eps_i a_i$ from \cite{MR1244666}, it follows that for $p=\log |H^\prime| = \gamma_2 N$,
\begin{align*}
\|Z_v-Z_w\|_{L_p} \gtrsim  & \max_{|I|=p} \Bigl( \sum_{i \in I} |v_i-w_i| + \sqrt{p} \bigl(\sum_{i \in  I^c} (v_i-w_i)^2\bigr)^{1/2} \bigr)
\\
\gtrsim  & \sqrt{p} \cdot \sqrt{\frac{c_1N}{8}} c_0 \|u_k - u_\ell\|_{L_2} \gtrsim \sqrt{c_1 \gamma_2} N \cdot c_0 \gamma_1 r.
\end{align*}
Indeed, for any subset of $\{1,...,N\}$ of cardinality $p \leq c_1N/8$, there are at least $c_1N/8$ coordinates in $I^c$ for which $|v_i-w_i| \geq c_0 \|u_k - u_\ell\|_{L_2}$; moreover, $\|u_k - u_\ell\|_{L_2} \geq \gamma_1 r$ because $H^\prime$ is $\gamma_1 r$-separated.

Hence, by Theorem \ref{thm:Sudakov-for-Bernoulli}, conditioned on the sample $(X_i)_{i=1}^N$,
$$
\E_\eps \sup_{u \in H_{h^*,r}} \left|\frac{1}{\sqrt{N}} \sum_{i=1}^N \eps_i u(X_i) \right| \gtrsim  c_0\sqrt{c_1} \cdot \gamma_1 \sqrt{\gamma_2}  N r,
$$
and since the `good event' has probability at least $1-2\exp(-c_1N/4) \geq 1/2$, it is evident that
$$
\E \sup_{u \in H_{h^*,r}} \left|\frac{1}{\sqrt{N}} \sum_{i=1}^N \eps_i u(X_i) \right| \gtrsim  c_0\sqrt{c_1} \cdot \gamma_1 \sqrt{\gamma_2}  N r,
$$
as required.
\endproof

\subsubsection*{Putting it all together}
Finally, all the ingredients are set in place to prove Claim \ref{claim:reduction}: that the combination of $\mP_1$ and $\mP_2$ leads to a $(1/20,r^\prime)$-essential subset of $H$ for $r^\prime \sim r$, which in turn leads to the wanted accuracy of $\eps$.

We continue by recalling the conditions on $\gamma_1,...,\gamma_4$ that have been collected along the way, and we may take the smallest values of the $\gamma_i$'s that emerge from all the conditions. Also, the fact that once a condition is verified for $N_0$ it is satisfied for any $N > N_0$ (with a modified constant) allows us to choose the smallest integer $N$ for which all the final conditions \eqref{eq:packing-condition}, \eqref{eq:Rad-Q-condition} and \eqref{eq:Rad-M-condition} are satisfied.

\vskip0.4cm

\subsubsection*{The distance oracle $\mP_1$}
Recall that we set
$$
\kappa_1=\max\{\kappa(1/10),1\}, \ \ \theta_0 \gtrsim \kappa_1^4, \ \ {\rm and} \ \ \ell = N/5\kappa_1^2.
$$
Let $N$ satisfy that
\begin{equation} \label{eq:final-cond-r-1}
\E \sup_{u \in H_{h^*,r}} \left|\frac{1}{\sqrt{N}} \sum_{i=1}^N \eps_i u(X_i) \right| \leq \frac{c_1}{\kappa_1^2} r \sqrt{N}, \ \ {\rm and} \ \ \log {\cal M} \left(H_{h^*,r}, \frac{c_2}{\kappa_1^2}D \right) \leq \frac{c_3}{\kappa_1^2}N
\end{equation}
for suitable absolute constants $c_1$, $c_2$ and $c_3$.

By Theorem \ref{thm:iso-two-sided}, there is an event ${\cal A}_1$ with probability at least
$$
1-2\exp(-c_4N/\kappa_1^4) \geq 1-\frac{\delta}{32}
$$
on which
$$
\frac{1}{2\sqrt{10}} \|h-h^*\|_{L_2} \leq \mP_1(h,h^*) \leq 3 \kappa_1 \|h-h^*\|_{L_2} \ \ {\rm if} \ \ \|h-h^*\|_{L_2} \geq r,
$$
and
$$
\mP_1(h,h^*) \leq 3\kappa_1 r \ \ {\rm if} \ \ \|h-h^*\|_{L_2} < r.
$$
Thus, using the notation of Section \ref{sec:det-to-rand}, we may set $\beta=3\kappa_1$ and $\alpha=1/2\sqrt{10}$; it is evident that Condition $(1)$ holds on ${\cal A}_1$.
\vskip0.3cm
Once the constants $\alpha$ and $\beta$ are specified, it forces the choice of $\nu$: following \eqref{eq:in-thm-cond-essential}, $\nu$ must satisfy that
$$
2\nu\left(\frac{\beta^2}{\alpha^2} +1 \right) \leq \frac{1}{20}.
$$
Therefore, we set
\begin{equation} \label{eq:final-cond-eps}
\nu \sim \frac{1}{\kappa_1^2}.
\end{equation}

\subsubsection*{An almost isometric lower bound}
Next, consider Condition $(2)$ from Section \ref{sec:det-to-rand}: the almost isometric lower bound on $\Q$, now with the constant $\nu$ set in \eqref{eq:final-cond-eps}. To verify the condition, let
$$
\xi_2=\frac{\nu}{4} \sim \frac{1}{\kappa_1^2}, \ \  \kappa_2=\kappa(\nu/4), \ \ \eta=0.01; \ \ \theta_0 \gtrsim \frac{\kappa_2^2}{\xi_2^2}; \ \ {\rm and} \ \ \theta_1 \gtrsim \frac{1}{\eta^2} \sim 1,
$$
and let $N$ satisfy that
\begin{equation} \label{eq:final-cond-r-2}
\E \sup_{u \in H_{h^*,r}} \left|\frac{1}{\sqrt{N}} \sum_{i=1}^N \eps_i u(X_i) \right| \leq \frac{c_5 \nu^2}{\kappa_2^2} r \sqrt{N} \ \ {\rm and} \ \ \log {\cal M} \left(H_{h^*,r}, \frac{c_6 \nu^2}{\kappa_2^2}rD \right) \leq \frac{c_7\nu^2}{\kappa_2^2}N
\end{equation}
for suitable absolute constants $c_5$, $c_6$ and $c_7$.

By Theorem \ref{thm:uniform-lower-isometric}, with probability at least $1-\delta/32$, if $\|h-h^*\|_{L_2} \geq r$ then on at least $0.98n$ of the coordinate blocks $I_j$,
$$
\Q_{h,h^*}(j) \geq (1-\nu)\|h-h^*\|_{L_2}^2,
$$
thus confirming Condition $(2)$ on that event.
\subsubsection*{A two-sided estimate on $\M$}
Finally, let use verify condition $(3)$ from Section \ref{sec:det-to-rand} --- the two-sided estimate on the multiplier component $\M$, again with the constant $\nu$ set in \eqref{eq:final-cond-eps}.

Let
$$
\eta=0.01; \ \ \theta_0 \gtrsim \frac{1}{\nu^2} \sim \kappa_1^4; \ \ {\rm and} \ \ \theta_1 \gtrsim 1,
$$
and let $N$ satisfy that
\begin{equation} \label{eq:final-cond-r-3}
\E \sup_{u \in H_{h^*,r}} \left|\frac{1}{\sqrt{N}} \sum_{i=1}^N \eps_i (h^*(X_i)-Y_i)u(X_i) \right| \leq c_8 \nu r^2 \sqrt{N}.
\end{equation}
for a suitable absolute constant $c_8$.

Applying Theorem \ref{thm:P3} for the levels $\rho=r $ and $\rho=(\beta/\alpha)r > r$, it follows that with probability at least $1-\delta/16$,
\begin{description}
\item{$\bullet$} if $\|h-h^*\|_{L_2} \geq r$ then
$$
|\M_{h,h^*}(j) - \E \M_{h,h^*} | \leq \nu \|h-h^*\|_{L_2}^2 \ \
{\rm on \ at \ least \ } 0.99n \ {\rm coordinate \ blocks};
$$
\item{$\bullet$} if $\|h-h^*\|_{L_2} \leq (\beta/\alpha)r$ then
$$
|\M_{h,h^*}(j) - \E \M_{h,h^*} | \leq \nu \frac{\beta^2}{\alpha^2} r^2 \ \ {\rm on \ at \ least \ } 0.99n \ {\rm coordinate \ blocks}.
$$
\end{description}
Moreover, since $\nu \beta^2/\alpha^2 \lesssim 1$ then $\gamma$ in Condition $(3)$ of Theorem \ref{thm:observation} is just an absolute constant.
\vskip0.3cm

Combining all these conditions, the constants needed for the definition of the $\mP$ are chosen to be
$$
\ell=\frac{N}{5 \kappa_1^2}; \ \ \theta_1 \sim 1; \ \ \theta_2 \sim \kappa_1^2; \ \ \theta_3 \sim \frac{1}{\kappa_1^2}; \ \ \theta_4 \sim \kappa_1.
$$
Also, the constants required for the sample complexity estimate are
$$
\theta_0 \sim \max\{\kappa_1^4,\kappa_1^2 \kappa_2^2\}
$$
and
$$
\gamma_1=\min\left\{\frac{c_2}{\kappa_1^2}; \frac{c_6 \nu^2}{\kappa_2^2} \right\} \ \ \ \gamma_2 = \min\left\{\frac{c_3}{\kappa_1^2}, \frac{c_7 \nu^2}{\kappa_2^2} \right\}; \ \ \ \gamma_3 = \min\left\{\frac{c_1}{\kappa_1^2},\frac{c_5 \nu^2}{\kappa_2^2}\right\}; \ \ \
\gamma_4=c_8\nu,
$$
where $\nu \sim 1/\kappa_1^2$. Thus, we require that
\begin{align*}
& \log {\cal M} \left(H_{h^*,r}, \gamma_1 rD \right) \leq \gamma_2 N; \ \ \ \E \sup_{u \in H_{h^*,r}} \left|\frac{1}{\sqrt{N}} \sum_{i=1}^N \eps_i u(X_i) \right| \leq \gamma_3 r \sqrt{N};
\\ {\rm and} \ \ & \E \sup_{u \in H_{h^*,r}} \left|\frac{1}{\sqrt{N}} \sum_{i=1}^N \eps_i (h^*(X_i)-Y_i) u(X_i) \right| \leq \gamma_4 r^2 \sqrt{N},
\end{align*}
and by Theorem \ref{thm:Bernoulli-implies-entropy}, the entropy condition is implied by an additional restriction on $\gamma_3$. Therefore, if we select $N$ as stated in Theorem \ref{thm:main} then all the three conditions hold.

With those set in place, it follows that with probability at least $1-\delta/2$,
the set $\mP_2(H)$ is a $(1/20,r^\prime)$-essential subset of $H$ where
$$
r^\prime=\sqrt{2}\left(\gamma+\frac{\beta^2}{\alpha^2}\right)^{1/2}r \sim \kappa_1 r;
$$
Setting $r \sim \eps/\kappa_1^2$ completes the proof of Claim \ref{claim:reduction} and therefore of Theorem \ref{thm:main} as well.
\endproof

\bibliographystyle{plain}
\bibliography{tournament3}

\newpage
\appendix

\section{Finite dictionaries revisited} \label{sec:dict}
Let $F=\{f_1,...,f_m\}$ be a finite dictionary, consider the triplet $(F,X,Y)$ and without loss of generality assume that $f^*=f_j$. To simplify notation, set $F_{j,r}= F_{f_j,r}$, and observe that it consists of functions of the form $f_\lambda=\lambda (f-f_j)$ where $0 \leq \lambda \leq 1$, $f \in F$ and $\|f_\lambda\|_{L_2} = \lambda \|f-f_j\|_{L_2} \leq r$. Therefore, if we set $u_{j,\ell}=(f_\ell-f_j)/\|f_\ell-f_j\|_{L_2}$ then
$$
F_{j,r} = \bigcup_{\ell=1}^m [0,r_{j,\ell} u_{j,\ell} ],
$$
where $0<r_{j,\ell} \leq r$ and $[0,h ]=\{\lambda h : 0 \leq \lambda \leq 1\}$.

The local geometry of $F$ is reflected in the structure of each $F_{j,r}$: these sets are the union of at most $m$ intervals $[0,v_i]$ for some $v_i \in L_2$, but the length of each interval and the `angles' between the intervals depend on $F$, $X$ and the specific centre $f_j$.
Note that the supremum $\sup_{u \in F_{j,r}} \left|\sum_{i=1}^N \eps_i u(X_i)\right|$ is attained at an extreme point of $F_{j,r}$ and those extreme points are $\{r_{j,\ell}u_{j,\ell} : 1 \leq \ell \leq m\}$. Therefore,
$$
(*)=\E \sup_{u \in F_{j,r}} \left|\frac{1}{\sqrt{N}} \sum_{i=1}^N \eps_i u(X_i)\right| =\E\max_{1 \leq \ell \leq m} \left|\frac{1}{\sqrt{N}}\sum_{i=1}^N \eps_i r_{j,\ell} u_{j,\ell}(X_i)\right|,
$$
and clearly the more heavy-tailed the random variables $u_{j,\ell}(X)$ are, the larger the expectation of the maximum of the $m$ random variables $Z_\ell=r_{j,\ell} |\sum_{i=1}^N \eps_i u_{j,\ell}(X_i)|$ will be.

It is natural to expect that the `extreme case' is when $r_{j,\ell}=r$ for every $\ell$ and the $u_{j,\ell}$ are orthogonal in $L_2$, that is, the localized set $F_{j,r}$ consists of $m$ orthogonal intervals of length $r$. However, even among these configurations there is still plenty of diversity: firstly, because orthogonality in $L_2$ of $(u_{j,\ell})_{\ell=1}^m$ does not imply that typical realizations of the vectors
$$
\left\{ \left(u_{j,\ell}(X_i)\right)_{i=1}^N : 1 \leq \ell \leq m \right\}
$$
are orthogonal or almost orthogonal in $\R^N$; and secondly, because the Bernoulli process $t \to \sum_{i=1}^m \eps_i t_i$ is not rotationally invariant, and expectations of normalized orthogonal configurations in $\R^N$ may differ by a factor of $\sqrt{\log N}$.

When the random variables $u_{j,\ell}(X)$ are orthogonal and well-behaved, estimating $(*)$ using the union bound is a reasonable strategy; in more general situations obtaining a sharp estimate on $(*)$ is significantly harder and requires sophisticated chaining methods.
\vskip0.3cm
This example illustrates the diversity one can expect---even in this simple learning scenario and for the same underlying distribution $X$ and target $Y$. It is a fact of life that $(*)$ may change substantially even among all classes consisting of the same number of points, and that is lost when considering only the worst case estimate.
\vskip0.3cm

A similar phenomenon occurs with the oscillation associated with the multiplier component: if we set $\xi_i=f_j(X_i)-Y_i$ then the oscillation is
$$
(**)= \E \sup_{u \in F_{j,r}} \left|\frac{1}{\sqrt{N}} \sum_{i=1}^N \eps_i (f_j(X_i)-Y_i) u(X_i)\right| = \E \max_{1 \leq \ell \leq m} \left|\frac{1}{\sqrt{N}}\sum_{i=1}^N \eps_i \xi_i r_{j,\ell} u_{j,\ell}(X_i)\right|;
$$
In other words, the $m$ random variables that one has to control are $Z_j=r_{j,\ell}|\sum_{i=1}^N \eps_i \xi_i  u_{j,\ell}(X_i)|$, with the additional complication of the multiplies $\xi_i$. $(**)$ captures the correlation between the vectors $(r_{j,\ell} u_{j,\ell}(X_i))_{i=1}^N$ and the noise vector $(\eps_i \xi_i)_{i=1}^N$, and even in relatively simple situations there is significant diversity in the geometry of the random set $\{(r_{j,\ell} u_{j,\ell}(X_i))_{i=1}^N : 1 \leq \ell \leq m\}$ and therefore in $(**)$.

\vskip0.5cm
\noindent {\bf Proof of \eqref{eq:worst-case-dictionary}.}
Recall that $F$ is a finite dictionary of cardinality $m$ and that for any $h \in {\rm span}(F)$ and every $p \geq 2$, $\|h\|_{L_p} \leq L \sqrt{p}\|h\|_{L_2}$.

Assume without loss of generality that $f^*=f_j=\inr{\cdot,t_j}$. Observe that for every $1 \leq \ell \leq m$, the random variables $(\eps_i u_{j,\ell}(X_i))_{i=1}^N$ are independent, mean-zero, variance $1$ and $L$-subgaussian. Therefore, each one of the random variables
$$
Z_\ell = \frac{1}{\sqrt{N}}\sum_{i=1}^N \eps_i r_{j,\ell} u_{j,\ell}(X_i)
$$
is mean-zero, has variance at most $r$ and is $cL$-subgaussian. By the union bound,
\begin{equation} \label{eq:dict-1}
\E\max_{1 \leq \ell \leq m} \left|\frac{1}{\sqrt{N}}\sum_{i=1}^N \eps_i r_{j,\ell} u_{j,\ell}(X_i)\right| \leq CL r \sqrt{\log m}.
\end{equation}
Next, consider the target $Y=\inr{t_0,X} + W$ for some $t_0 \in \R^d$ and $W$ that is square integrable and independent of $X$. Note that
\begin{align*}
& \E \max_{1 \leq \ell \leq m}\left|\frac{1}{\sqrt{N}}\sum_{i=1}^N \eps_i \xi_i r_{j,\ell} u_{j,\ell}(X_i)\right|
\\
\leq & \E \max_{1 \leq \ell \leq m} \left|\frac{1}{\sqrt{N}}\sum_{i=1}^N \eps_i \inr{t_j-t_0,X_i} \cdot r_{j,\ell} u_{j,\ell}(X_i)\right| + \E \max_{1 \leq \ell \leq m} \left|\frac{1}{\sqrt{N}}\sum_{i=1}^N \eps_i W_i r_{j,\ell} u_{j,\ell}(X_i)\right| = (1)+(2).
\end{align*}
An estimate for $(1)$ is a straightforward outcome of Corollary 1.10 from \cite{Men-multi16}, showing that
$$
(1) \lesssim L^2 \|\inr{t_j-t_0,X}\|_{L_2} r \sqrt{\log m}.
$$
Also, by Lemma 2.9.1 from \cite{vaWe96} followed by the union bound,
$$
\E \max_{1 \leq \ell \leq m} \left|\frac{1}{\sqrt{N}}\sum_{i=1}^N \eps_i W_i r_{j,\ell} u_{j,\ell}(X_i)\right| \lesssim L \|W\|_{L_2} r \sqrt{\log m}.
$$
Hence,
\begin{equation} \label{eq:dict-2}
\E \max_{1 \leq \ell \leq m}\left|\frac{1}{\sqrt{N}}\sum_{i=1}^N \eps_i \xi_i r_{j,\ell} u_{j,\ell}(X_i)\right| \leq c(L) \sigma r \sqrt{\log m},
\end{equation}
and the claim follows from \eqref{eq:dict-1} and \eqref{eq:dict-2}.
\endproof

\section{The bounded framework} \label{sec:bounded}
The bounded framework focuses on triplets $(F,X,Y)$ where $F$ consists of functions that are bounded by $M$ and $Y$ is also bounded by $M$ (often one sets $M=1$). The obvious downside of this framework is its limited scope. Indeed, the simplest of statistical models, independent additive gaussian noise (i.e., $Y =f_0(X)+W$ for $f_0 \in F$ and a fixed, centred gaussian random variable $W$) is out of the bounded framework's reach even if $f_0$ is bounded by $M$. However, being the natural extension of binary classification, and because the technical machinery required for its study was available, the bounded framework has been of central importance since the very early days of statistical learning theory.

It is natural to expect that addressing problems that belong to the bounded framework would be easier than the general, heavy-tailed scenario studied here. Indeed, thanks to Talagrand's celebrated concentration inequality for bounded empirical processes \cite{MR1258865} (see also \cite{BoLuMa13}), at the heart of the work on the bounded framework is the fact that empirical means exhibit strong, uniform concentration around the true means. 

Thanks to the powerful technical machinery that has been available for bounded problems, there has been progress in the study of unrestricted procedures in that setup. For example, optimal worst-case estimates were obtained in \cite{LecMenAgg09} for finite dictionaries and in \cite{MR3606751} for classes that satisfy a uniform random entropy condition. However, despite this progress the picture was far from complete: there was no `unrestricted analog' of the known estimates for convex classes in the bounded framework. The state of the art estimates in that case were obtained in \cite{BaBoMe05} (see also \cite{Kolt08}), where the following was shown:
\begin{Theorem} \label{thm:bounded-convex}
There are absolute constant $c_1$ and $c_2$ for which the following holds. Let $F$ be a convex class consisting of functions bounded by $M$ and assume that $Y$ is also bounded by $M$. If $r$ is such that
$$
\E \sup_{u \in F_{f^*,r}} \left|\frac{1}{N} \sum_{i=1}^N \eps_i  u(X_i) \right| \leq c_1 \frac{r^2}{M},
$$
then with probability at least $1-2\exp(-t)$, given a sample $(X_i,Y_i)_{i=1}^N$, ERM selects $\hat{f}$ that satisfies
$$
\|\hat{f}-f^*\|_{L_2}^2 \leq c_2\max\left\{r^2,t\frac{M^2}{N}\right\}.
$$
\end{Theorem}
In other words, if we set $r^2=\eps/c_2$ and $N \gtrsim \frac{M^2}{\eps} \log(2/\delta)$ then with probability at least $1-\delta$, $\|\hat{f}-f^*\|_{L_2} < \eps$. 

\vskip0.4cm
We will show that for an appropriate modification of $\mP_1$, using $\mP$ leads to an unrestricted analog of Theorem \ref{thm:bounded-convex}, and the proof follows an identical path to the proof of Theorem \ref{thm:main}.

The modification in $\mP_1$ is unavoidable: a bounded function $w$ need not assign any weight to an interval $[c_1\|w\|_{L_2},c_2\|w\|_{L_2}]$; therefore, selecting $(w(X_i))_\ell^*$ as an estimator of $\|w\|_{L_2}$ is a poor choice. On the other hand, the most trivial estimator of distances is good enough. Indeed, let $H$ be a class of functions that are bounded by $M$. A standard application of Talagrand's concentration inequality followed by symmetrization and contraction shows that if
$$
\E \sup_{u \in H_{h^*,r}} \left|\frac{1}{N} \sum_{i=1}^N \eps_i  u(X_i) \right| \leq C \frac{r^2}{M} 
$$
then with probability at least $1-2\exp(-cN\min\{r^2/M^2,1\})$, for any $h \in H$ such that $\|h-h^*\|_{L_2} \geq r$, one has
\begin{equation} \label{eq:norm-equiv-sec-4}
\frac{1}{2}\|h-h^*\|_{L_2} \leq \Bigl(\frac{1}{N}\sum_{i=1}^N (h-h^*)^2(X_i) \Bigr)^{1/2} \leq 2 \|h-h^*\|_{L_2}.
\end{equation}
This provides sufficient information for the success of $\mP_2$, and $\mP_1$ may be replaced by the empirical $L_2$ distance between any two functions in the class. 

The second part of the procedure, $\mP_2$ remains unchanged (though the tuning parameters have be adapted to the change in $\mP_1$). It is straightforward to see that the tuning parameters turn out to be just absolute constants.

To formulated a version of Theorem \ref{thm:main} in the bounded framework, let $T=(H,X,Y)$ be a triplet where $H$ consists of functions bounded by $M$ and $Y$ is also bounded by $M$. Set
$$
N_3(H,r,\kappa) = \min\left\{ N : \E \sup_{u \in H_{h^*,r}} \left|\frac{1}{N} \sum_{i=1}^N \eps_i  u(X_i) \right| \leq \kappa \frac{r^2}{M}  \right\}.
$$
\begin{Theorem} \label{thm:bounded-framework}
There exist absolute constants $c_0,c_1$ and $c_2$ for which the following holds. Given $\eps$ and $\delta$, let $r^2=c_0 \eps$ and set 
$$
N \geq 2\left(N_3(F,r,c_1) + N_3(\bar{F}_1,r,c_1)\right) + c_2 \frac{M^2}{\eps} \log\left(\frac{64}{\delta}\right).
$$
Then
$$
\E \left((\tilde{f}(X)-Y)^2 | (X_i,Y_i)_{i=1}^{N}\right) \leq \inf_{f \in F} \E (f(X)-Y)^2 + \eps \ \ \ {\rm with \ probability \ } 1-\delta. 
$$
\end{Theorem}

It is important to stress that Theorem \ref{thm:bounded-framework} is a worst-case estimate. While it does capture some of the interplay between $F$ and $X$ (reflected by the condition on the Rademacher averages), it is attained by eliminating the `noise level' $\sigma$ from the problem. As a result, the estimate is suboptimal when $\sigma$ is small and one has more information other than just knowing that the class and target are bounded by $M$. Theorem \ref{thm:bounded-framework} does not `see' the full diversity displayed by triplets, whereas Theorem \ref{thm:main} does.

\vskip0.4cm

All the standard sample complexity estimates in the bounded framework can be recovered from Theorem \ref{thm:bounded-framework}, simply by establishing upper estimates on the localized Rademacher averages appearing in the definition of $N_3$. For example, it is straightforward to verify that way that if $F$ is a finite dictionary of cardinality $m$ then it suffices that
$$
N \gtrsim  \left(\frac{M^2}{\eps}+1\right) \cdot \left(\log m + \left(\frac{64}{\delta}\right) \right);
$$
this recovers the main result from \cite{LecMenAgg09}. A similar argument may be used to recover the results from \cite{MR3606751}, by using the given information on uniform entropy numbers to control the localized Rademacher averages.

Of course, the procedure we use here is different from the one normally employed in the bounded framework: $\mP$ is not ERM based.

\vskip0.4cm

The components of the proof of Theorem \ref{thm:bounded-framework} are identical to those of Theorem \ref{thm:main}, while keeping track of the modified constants because of the altered $\mP_1$. Let us sketch the minor differences:

\begin{description}
\item{$\bullet$} Using the notation of Section \ref{sec:det-to-rand}, the modified $\mP_1$, defined by 
$$
\mP_1^2(h,f)=\frac{1}{N}\sum_{i=1}^N (h-f)^2(X_i),
$$
satisfies Condition $(1)$ for absolute constants $\alpha$ and $\beta$.  
\item{$\bullet$} The first part of Condition $(2)$ is proved in \cite{Men:gen} (see the `moreover' part of Corollary 3.6 there).
\item{$\bullet$} The second component of Condition $(2)$ and Condition $(3)$ are outcomes of Theorem \ref{thm:P3}. Note that its proof does not use the uniform integrability condition at all, and thus may be applied directly, without changes. Clearly, if $F$ consists of functions bounded by $M$ then one may take $L_T = M$ and $\sigma \leq 2M$. 
\item{$\bullet$} The required version of Theorem \ref{thm:Bernoulli-implies-entropy} (actually, a more general version) can be found in \cite{Men:gen}.
\end{description}

\end{document}